# MapGlue: Multimodal Remote Sensing Image Matching

Peihao Wu, Yongxiang Yao*, Wenfei Zhang, Dong Wei, Yi Wan, Yansheng Li, Yongjun Zhang*

*Abstract*—Multimodal remote sensing image (MRSI) matching is pivotal for cross-modal fusion, localization, and object detection, but it faces severe challenges due to geometric, radiometric, and viewpoint discrepancies across imaging modalities. Existing unimodal datasets lack scale and diversity, limiting deep learning solutions. This paper proposes MapGlue, a universal MRSI matching framework, and MapData, a large-scale multimodal dataset addressing these gaps. Our contributions are twofold. MapData, a globally diverse dataset spanning 233 sampling points, offers original images (7,000×5,000 to 20,000×15,000 pixels). After rigorous cleaning, it provides 121,781 aligned electronic map–visible image pairs (512×512 pixels) with hybrid manual-automated ground truth, addressing the scarcity of scalable multimodal benchmarks. MapGlue integrates semantic context with a dual graph-guided mechanism to extract cross-modal invariant features. This structure enables global-to-local interaction, enhancing descriptor robustness against modality-specific distortions. Extensive evaluations on MapData and five public datasets demonstrate MapGlue's superiority in matching accuracy under complex conditions, outperforming state-of-the-art methods. Notably, MapGlue generalizes effectively to unseen modalities without retraining, highlighting its adaptability. This work addresses longstanding challenges in MRSI matching by combining scalable dataset construction with a robust, semantics-driven framework. Furthermore, MapGlue shows strong generalization capabilities on other modality matching tasks for which it was not specifically trained. The dataset and code are available at https://github.com/PeihaoWu/MapGlue.

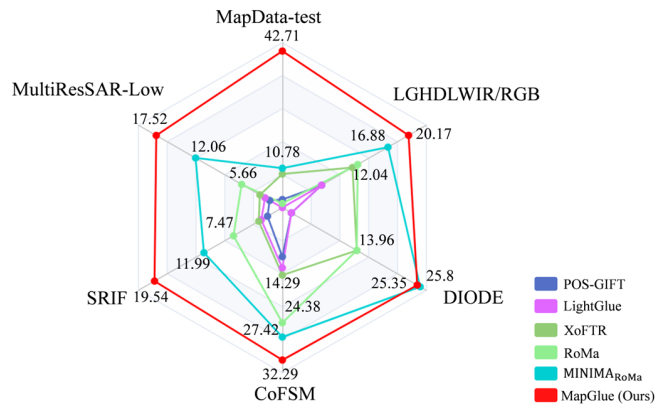

Fig. 1. Overall Matching Accuracy on Six Multimodal Datasets. The average AUC under a reprojection error (@5px) threshold is used for accuracy evaluation. Compared with traditional methods and representative algorithms for sparse, semi-dense, and dense matching, our MapGlue method exhibits exceptional stability and generalization across multiple datasets.

## I. INTRODUCTION

Multimodal remote sensing image (MRSI) matching, as a key technology for cross-modal data alignment, aims to establish pixel-level or feature-level correspondences among heterogeneous imaging data (e.g., visible, SAR, thermal infrared) [1,19,20], thereby achieving semantic alignment and geometric registration across domains. In recent years, with the rapid development of multisensor fusion technologies [1], this research direction has made groundbreaking progress at the intersection of computer vision and remote sensing. Its core value lies not only in enhancing the consistency of cross-modal data representation, but also in overcoming the inherent limitations of individual sensors, such as illumination adaptability, texture sensitivity, and geometric invariance, through modality complementarity. In critical applications including high-precision positioning for autonomous driving, urban 3D reconstruction, and disaster emergency response, MRSI matching has become a cornerstone for robust environmental perception [2,3,4]. Particularly in the field of electronic navigation maps, cross-modal matching between electronic navigation maps and visible light images, by fusing visual semantic features with topological road network structures, provides a new paradigm for high-precision crowdsourced map updates and real-time positioning. However, existing methods still encounter dual bottlenecks in both theoretical modeling and engineering practice when addressing challenges such as non-linear radiometric differences (NRD), non-rigid geometric distortions (NGD), and cross-domain representation gaps [5,20,28,29,30,59].

Currently, the primary limitation in MRSI matching research is the scarcity of high-quality benchmark datasets [6,7,19,54]. This dilemma arises from two fundamental contradictions. First, the temporal, spatial, and spectral differences among various sensors lead to an exponential increase in the cost of acquiring high-quality, real multimodal paired data. Second, the annotation process must satisfy both pixel-level geometric precision and semantic consistency, making traditional annotation approaches inadequate.

This work was supported in part by the National Key Research and Development Program of China, Grant No. 2024YFB3909001, the Key Program of the National Natural Science Foundation of China, Grant No. 42030102, the Program of the National Natural Science Foundation of China, Grant No. 42401534, No. 42301499, and No. 42471470. *(Corresponding author: Yongxiang Yao and Yongjun Zhang).*

All authors are with the School of Remote Sensing Information Engineering, Wuhan University, Wuhan 430079, China (e-mail: (wupeihao, yaoyongxiang, zhangwenfei, weidong, yi.wan, yansheng.li, zhangyj)@whu.edu.cn).

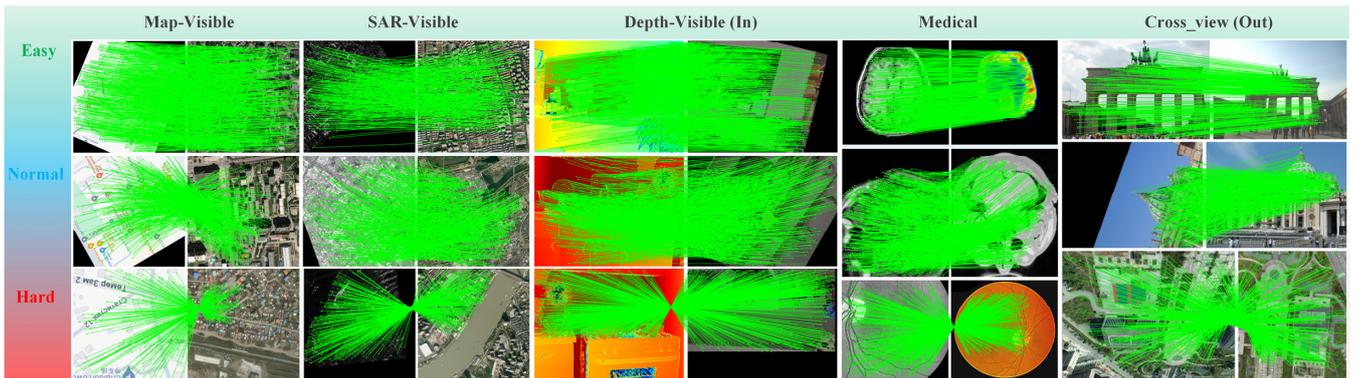

Fig. 2. Qualitative Results of Multimodal Image Matching. "Easy," "Normal," and "Hard" denote different levels of transformations applied to the images.

Accurate ground truth labeling is not only time-consuming but also costly, which further restricts the scale and availability of datasets. Existing solutions often rely on transfer learning from homogeneous visible light image datasets or on generating simulated data based on physical imaging models [6,54]. However, such methods suffer from significant domain shift issues. For example, simulated data in SAR-visible image matching often fail to capture the complex spatial distribution of radar cross-section (RCS). In map–visible image matching, inherent differences between the topological abstraction of vector data and the detailed texture of raster images create a modality gap between simulated and real data, significantly degrading the performance of deep matching models in practical applications.

To systematically address these issues, we have mobilized a substantial research team to construct a large-scale, multi-scene, and multimodal electronic navigation map–visible light image dataset, referred to as MapData, specifically for electronic navigation map applications. The goal of MapData is to provide high-quality training and evaluation resources for global-scale remote sensing image matching. Unlike natural image matching, MRSIs typically exhibit significant geometric distortions due to differences in imaging mechanisms. This necessitates that matching algorithms not only achieve precise local correspondences but also perform effective global alignment to prevent the accumulation of global matching errors. To this end, we propose MapGlue, a cross-modal matching method based on a dual graph neural network that integrates semantic information. Specifically, in the keypoint extraction stage, we introduce a saliency feature map constraint to ensure that the extracted keypoints are uniformly distributed in salient regions. Next, for feature descriptors, we design a saliency feature descriptor that incorporates image semantic information to comprehensively characterize both structural and textural features. Finally, in feature matching, we propose a dual graph structure network to guide the allocation of attention during feature interaction, enabling a gradual transition from capturing global structural features to local detailed features, thereby synergistically enhancing both global and local representations.

Comparative experiments with mainstream methods validate the effectiveness of the MapData dataset and demonstrate the superior performance of the MapGlue method. The quantitative matching accuracy results across six multimodal datasets are shown in Fig. 1, while the qualitative matching results of MapGlue on multimodal images are shown in Fig. 2. Specifically, our main contributions are as follows:

(1) We introduce MapData, a large-scale, multi-scene dataset for map–visible image cross-modal registration, which provides abundant training and evaluation resources for the development of MRSI matching algorithms.

(2) We propose MapGlue, an efficient and robust cross-modal matching method based on a dual graph neural network that integrates semantic information. By extracting salient keypoints, fusing semantic information from images, and constructing a dual graph structure for guided enhancement, matching accuracy and robustness are significantly improved.

(3) Extensive comparisons with existing methods on six multimodal datasets validate our approach: both quantitative and qualitative experiments demonstrate the superior performance of MapGlue in multimodal matching tasks. Furthermore, the results show that MapGlue exhibits strong generalization capabilities, maintaining stability and applicability in complex cross-modal matching scenarios without additional training.

## II. RELATED WORK

In this section, we first discuss the datasets used for image matching, and then review the conventional handcrafted methods and deep learning methods employed in image matching.

### A. Image Matching Datasets

As shown in Table I, existing image matching datasets can be categorized into homologous and multimodal based on modality homogeneity, with significant differences in their data distributions and annotation characteristics.

*1) Homologous Image Matching Datasets.* In the realm of homologous matching, researchers have constructed several datasets. Among these, the most classic and widely used is MegaDepth [8], a large-scale outdoor RGB image matching dataset comprising approximately 40 million image pairs. In this dataset, the images vary in size, with an average resolution of approximately 1280 × 960 pixels, and subpixel-level camera pose ground truth is generated using Structure-from-Motion (SFM) algorithms [9]. The ScanNet [10] dataset focuses on indoor scenes and provides 230 million RGB image pairs at 1296×968 pixels, with its annotation system integrating 3D camera poses, surface mesh reconstructions, and instance-level semantic segmentation to form a multi-task benchmark. HPatches [11] is a multi-view sequence dataset that contains 15 groups of high-density image sequences across 116 scenes, with each sequence composed of hundreds to thousands of 65 × 65 pixels image patches. The entire dataset comprises 2 million pairs of 65 × 65 pixels image patches, each accompanied by precisely annotated homography matrices.

TABLE I. Overview of MapData and Other Matching Datasets. (RGB indicates visible images, IR indicates infrared images, Out indicates outdoor scenes, In indicates indoor scenes, RS indicates remote sensing, and Modality denotes the number of modalities.)

| Dataset | Number of pairs | Image size | Scene | Modality | Label |
|---|---|---|---|---|---|
| **Homologous image matching dataset** | | | | | |
| MegaDepth [8] | 40M | 1280 × 960 | Out | 1 | Depth,Pose |
| ScanNet [10] | 230M | 1296 × 968 | In | 1 | Depth,Pose |
| HPatches [11] | 2M | 65 × 65 | Out | 1 | Homography |
| **Multimodal image matching dataset** | | | | | |
| LGHD LWIR/RGB [13] | 44 | 639 × 431 | Out | 2 | Homography |
| LLVIP [14] | 17K | 1280 × 1024 | Out | 2 | Homography |
| KAIST Multispectral Pedestrian Dataset [15] | 95K | 640 × 480 | Out | 2 | Homography |
| RGB-NIR Scene [16] | 477 | 1024 × 768 | Out & in | 2 | Homography |
| DIODE [17] | 25K | 1024 × 768 | Out & in | 3 | Homography |
| METU-VisTIR [18] | 2.5K | 3840 × 2160 (RGB)  640 × 512 (IR) | Out | 2 | Pose |
| SRIF Dataset [19] | 1.2K | 256 × 256 | RS | 7 | Homography |
| CoFSM Dataset [20] | 60 | 400 × 400 ~ 661 × 661 | RS | 7 | Homography |
| GoogleMap [21] | 9.6K | 192 × 192 | RS | 2 | Homography |
| MultiResSAR [22] | 10K | 512 × 512 | RS | 2 | Homography |
| **MapData (ours)** | **120K** | **512 × 512** | **RS** | **2** | **Homography** |

The advantage of homologous datasets lies in the feasibility of data acquisition and annotation: dense correspondences can be automatically generated via 3D reconstruction techniques (e.g., COLMAP) [12], and the ubiquity of visible light sensors enables large-scale scene coverage. However, since these datasets cannot model the non-linear mapping relationships between different sensors, they face severe domain adaptation challenges in cross-modal matching tasks.

*2) Multimodal Image Matching Datasets.* To overcome the modality gap, researchers have sequentially developed several small-scale multimodal image matching datasets. These datasets cover various modality combinations, including infrared-visible, SAR-visible, depth-visible, map–visible, as well as multi-temporal/day-night modalities.

In the natural image domain, Aguilera et al. introduced the LGHD LWIR/RGB [13] dataset, which contains 44 pairs of aligned 639×431 pixels infrared-visible images. Jia et al. proposed the LLVIP [14] dataset, comprising 17K pairs of aligned 1280 × 1024 pixels outdoor infrared–low visible images. Hwang et al. presented the KAIST Multispectral Pedestrian Dataset [15], which includes 95K pairs of aligned 640 × 480 pixels thermal-visible images covering vehicles and pedestrians. Brown et al. proposed the RGB-NIR Scene [16] dataset, which consists of 477 pairs of aligned 1024 × 768 pixels near-infrared and visible light images across 9 categories. Vasiljevic et al. presented the DIODE [17] dataset, which consists of 25K pairs of aligned 1024 × 768 pixels indoor and outdoor depth–visible images. Tuzcuoğlu et al. introduced the METU-VisTIR [18] dataset, containing 2.5K pairs of outdoor infrared-visible images, where the visible light images are 3840 × 2160 pixels and the infrared images are 640 × 512 pixels, with camera pose information provided as matching labels.

In the remote sensing domain, Li et al. introduced the SRIF [19] dataset, which includes seven types of modality data. Each of the remaining modalities is paired with visible light images to form six distinct categories of datasets (e.g., multi-temporal–visible, infrared–visible, depth–visible, map–visible, SAR–visible, and day-night images). Each modality consists of 200 pairs of data, totaling 1.2K pairs, with an image resolution of 256 × 256 pixels and matching ground truth provided in the form of affine transformation matrices. Yao et al. proposed the CoFSM [20] dataset, which also includes seven types of modality images including visible light, with each modality containing 10 pairs of images (totaling 60 pairs) and image size ranging from 400 to 661 pixels; the matching ground truth is provided as corresponding keypoint pairs. Zhao et al. introduced the GoogleMap [21] dataset, which contains 9600 pairs of aligned map–visible images with a resolution of 192 × 192 pixels. However, due to its relatively low resolution, its applicability is limited for training high-precision registration models. Zhang et al. presented the MultiResSAR [22] dataset, comprising over 10K pairs of multi-source, multi-resolution, and multi-scene SAR-visible images, each of size 512 × 512 pixels, with the matching ground truth provided as homography matrices.

Although the aforementioned multimodal datasets have provided valuable data resources for related research, they generally suffer from small scales and limited scene diversity. This is especially true in the remote sensing domain, where there is a lack of sufficiently large-scale datasets to meet the massive data requirements for deep learning training. Therefore, our constructed large-scale electronic navigation map–visible light image dataset, MapData, will offer robust data support for training deep matching models and drive the advancement of MRSI matching research.

*B. Image Matching Approaches*

Image matching is a core task in computer vision, with wide-ranging applications including camera pose estimation, geolocation, image fusion, and object detection. In this domain, matching methods can be broadly categorized into two groups: traditional handcrafted methods and modern deep learning–based methods [60]. In this subsection, we discuss the development and applications of image matching techniques from both perspectives.

*1) Traditional Matching Methods.* Early image matching methods primarily relied on the construction of keypoints and descriptors, with the most classical and widely used techniques being SIFT [23] and SURF [24], which employ scale-invariant feature transforms for matching. To overcome NRD and NGD in multimodal imagery, subsequent research has made significant improvements in keypoint extraction and descriptor construction. For example, Ma et al. proposed the position-scale-orientation SIFT (PSO-SIFT) [25] to address non-linear brightness variations and rotational changes. Xiong et al. introduced the oriented self-similarity (OSS) [26] and adjacent self-similarity (ASS) [27] methods, which are tailored for handling modality differences and rotational variations in MRSIs, demonstrating stability in scenarios with minor NRD. For images with severe radiometric distortions, Li et al. proposed RIFT [28] and RIFT2 [29]. While RIFT utilizes a polar coordinate grid to design the descriptor and reduces dimensionality to enhance computational efficiency, its performance slightly deteriorates in the presence of noise, illumination, and scale changes. RIFT2 further improves robustness, especially for remote sensing or complex texture images. Yao et al. introduced the matching method based on histogram of absolute phase consistency gradients (HAPCG), along with the multi-orientation tensor index feature (MoTIF) [31] and the histogram of the orientation of weighted phase (HOWP) [32] methods, which can effectively overcome differences in scale, translation, and rotation. Among these, HOWP exhibits superior performance in matching MRSIs with significant noise differences; however, the stability of these three methods remains limited when facing large rotational variations. Recently, Liao et al. proposed a repeatable feature refinement method [33] that reduces the impact of modality differences on matching by extracting repeatable features from multimodal images. They also introduced the AMSE [34] method, which adaptively adjusts filter parameters to reduce the complexity of manual tuning, thereby improving matching success rate and precision. Hou et al. proposed POS-GIFT [35] method, which employs a multi-layer circular sampling strategy, a rotation-invariant descriptor, and a position–orientation–scale guided inlier recovery strategy, achieving promising results on multimodal data. However, despite continuous improvements in addressing radiometric, illumination, and geometric differences in multimodal imagery, traditional methods still suffer from a limited capacity to represent multimodal features. Consequently, they are still unable to solve the comprehensive matching challenges in cross-scale, cross-rotation, cross-view, and cross-modal complex scenes.

*2) Deep Learning–Based Matching Methods.* With the rapid development of deep learning, these techniques have been widely applied to image matching tasks. Deep learning matching methods can be roughly divided into three categories: detector-based methods, detector-free methods, and other matching methods.

*Detector-Based Matching Methods.* In detector-based methods, examples such as MagicPoint [36], SuperPoint [37], and R2D2 [38] utilize self-supervised training for keypoint extraction, thereby overcoming the limitations of traditional keypoint detectors. D2-Net [39] extracts both keypoints and descriptors simultaneously over the entire image, significantly improving detection efficiency. Building upon these methods, SuperGlue [40] and LightGlue [41] enhance the descriptors extracted by SuperPoint [37] using graph neural networks, thereby boosting both matching accuracy and algorithmic efficiency. OmniGlue [42] further incorporates information extracted by DINOv2 [43] to guide the feature propagation process between images, encouraging the model to exploit potential matching regions, which notably enhances its generalization across multiple datasets. XFeat [44] achieves higher speed through a lightweight keypoint and descriptor extraction network, although its generalization capability across different datasets is somewhat limited. DKDNet [45] introduces a dynamic keypoint detection network based on attention mechanisms, significantly improving the efficacy of keypoints and matching accuracy. Overall, detector-based matching methods rely on the precision of keypoint and descriptor extraction, with the effective distribution of keypoints and the richness of descriptor information directly impacting the final matching outcomes.

*Detector-Free Matching Methods.* Detector-free methods bypass the keypoint extraction stage entirely, leveraging the image feature maps directly for matching. For instance, LoFTR [46] pioneered a semi-dense matching approach, progressing from coarse to fine matching and achieving subpixel-level precision, thus providing an important reference for subsequent semi-dense matching research. Building on LoFTR [46], ELoFTR [47] improves the attention mechanism by proposing an aggregated attention module, which not only optimizes computational efficiency but also enhances matching accuracy. XoFTR [18], tailored for RGB-IR image matching, uses the MAE mask training method [48], pretraining on the 95K pairs of visible–thermal infrared images from the KAIST Multispectral Pedestrian Detection Dataset [15] and fine-tuning on infrared images to achieve effective multimodal matching. DKM [49] and RoMa [50] are dense matching methods that can generate a large number of matching points, resulting in significant improvements in matching metrics. However, these methods also have higher computational costs and are less efficient. Additionally, analysis from existing studies reveals that detector-free methods may tend to concentrate matching points in local regions when processing images with high matching difficulty, thereby falling into local optima. Given that MRSIs often exhibit global distortions, detector-free methods may display unstable performance in multimodal matching tasks.

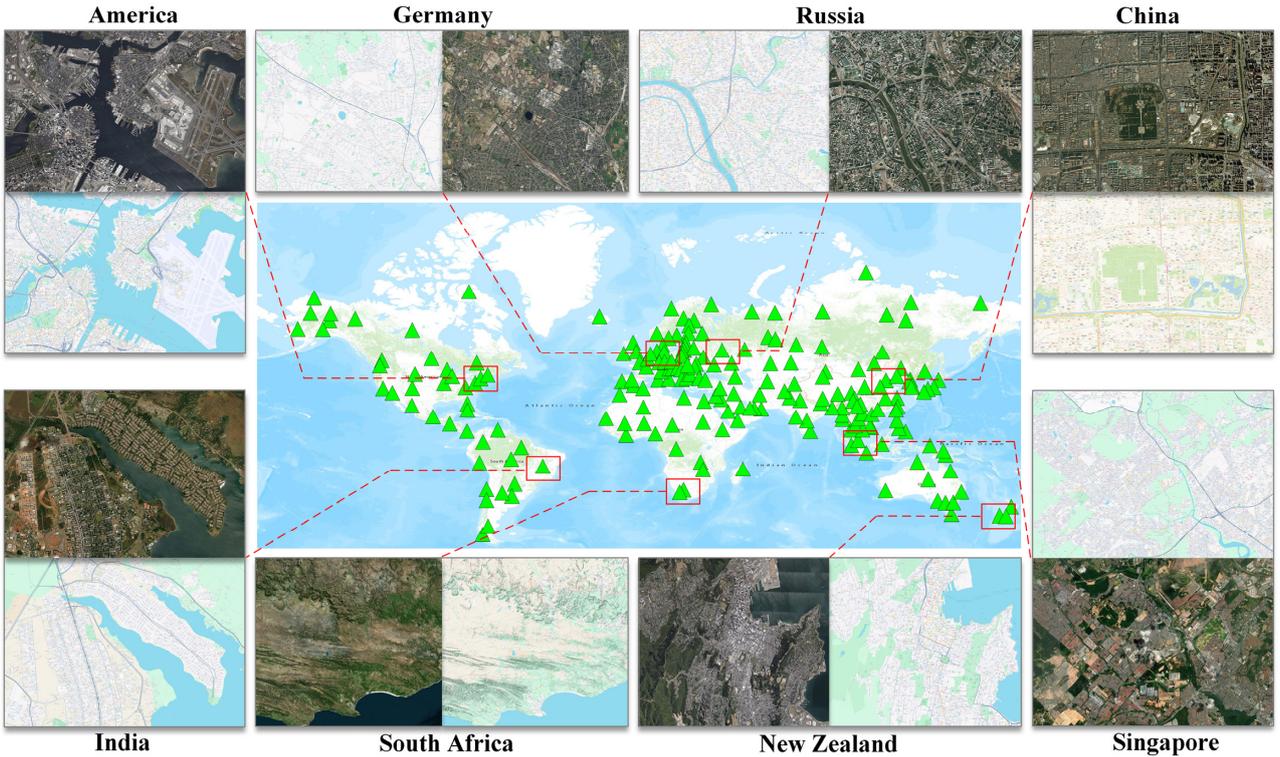

Fig. 3. Geographic Distribution of Sampled Images in the MapData Dataset.

*Other Matching Methods.* Beyond the aforementioned approaches, several other methods exist in the field of deep learning–based registration. For example, MCNet [51] combines multi-scale feature extraction with a correlation search strategy to progressively optimize image correlation at multiple resolutions, directly estimating the homography transformation between images. SuperRetina [52] is designed specifically for retinal image matching, employing an end-to-end framework to jointly train keypoint detectors and descriptors using partially annotated retinal image pairs. It also uses a progressive keypoint expansion strategy in a semi-supervised manner. GIM [53] uses consecutive video frames to train networks such as LightGlue [41], LoFTR [46], and DKM [49], resulting in models ($GIM_{LG}$, $GIM_{LoFTR}$, and $GIM_{DKM}$) with enhanced generalization on homologous images. Furthermore, $MINIMA_{LG}$, $MINIMA_{LoFTR}$, and $MINIMA_{RoMa}$ [54] are trained on simulated multimodal image matching datasets, which significantly improve cross-modal matching generalization performance, achieving promising results on several cross-modal datasets and demonstrating the critical role of large-scale data samples in deep learning methods.

In summary, although both traditional and existing deep learning matching methods have made progress in multimodal image matching, they still face the comprehensive challenge of overcoming variations in scale, rotation, viewpoint, and modality in complex scenes. Particularly, models trained on natural and simulated images often suffer from domain shift when generalized to actual remote sensing images, which fundamentally differ in texture distribution and geometric deformation, leading to degraded performance. To address these issues, we have constructed a large-scale MapData dataset to enrich the MRSI matching resources and designed a network model specifically tailored for MRSI matching. This approach significantly improves the stability and accuracy of MRSI matching, demonstrating immense potential for practical engineering applications.

## III. MAPDATA DATASET

Image collection for the MapData dataset began in August 2023 and labeling was completed in May 2024, a total of 10 months and more than 20 people were involved.

### A. Data Collection

This dataset was constructed by obtaining electronic navigation map data using the Google Maps API and downloading corresponding visible light images through the Google Earth Engine (GEE) platform, thereby creating the multimodal image dataset MapData. Data collection covers

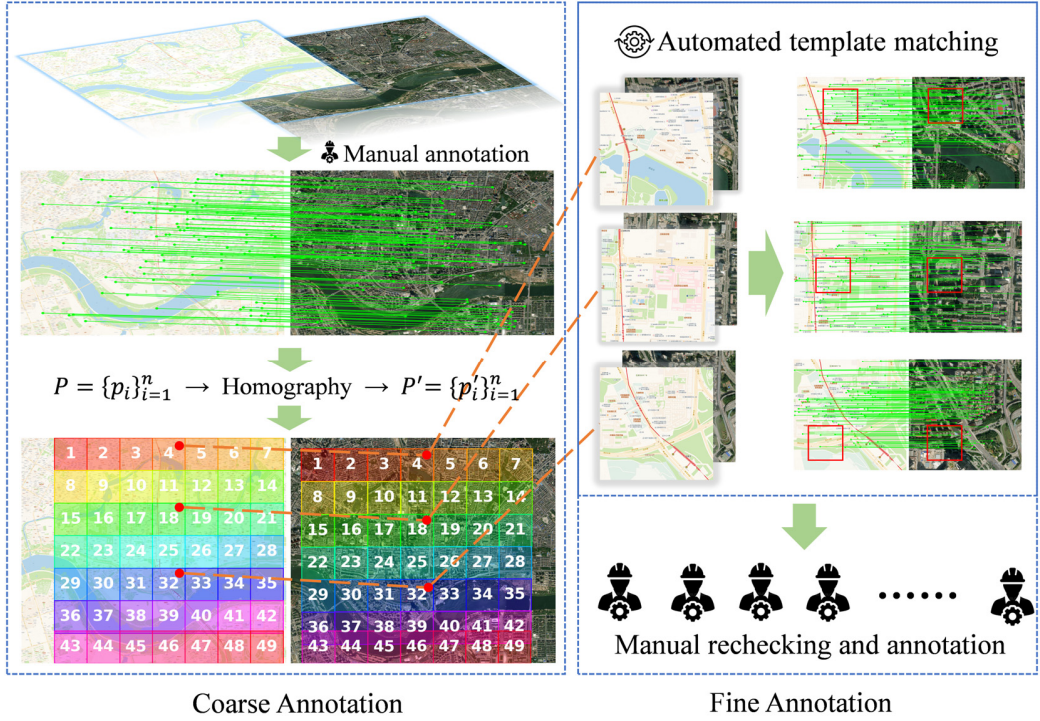

Fig. 4. Annotation Process Flowchart. In the coarse stage, images are manually annotated and divided into small patches; in the fine stage, automated annotation is combined with manual re-inspection.

233 geographic sampling points worldwide, encompassing a variety of typical scenes such as urban built-up areas, rural settlements, mountainous terrain, plains and basins, and desert regions. The spatial resolution of the original imagery ranges from 3 to 500 meters, with pixel dimensions varying from 7000 × 5000 to 20000 × 15000. A total of 170,162 image pairs were initially generated, and after rigorous quality screening, 121,781 valid pairs were retained. The dataset was randomly partitioned into training, validation, and test sets, containing 109,871 pairs, 10,000 pairs, and 1,910 pairs, respectively, while ensuring balanced geographic distribution and scene diversity. The global spatial distribution of the dataset and visual comparisons of representative samples are shown in Fig. 3.

*B. Data Annotation*

To address the cross-modal annotation challenge between electronic navigation maps and visible light images, as shown in Fig. 4, we designed a two-stage hybrid annotation framework.

*1) Coarse Annotation Stage.* First, 200 uniformly distributed reference points were manually annotated on the large-scale images. An initial homography matrix $H$ was then estimated using the RANSAC algorithm, ensuring that the reprojection error was controlled within 3 pixels. Next, a block partitioning strategy was employed to divide the original images into sub-images of 512 × 512 pixels. To avoid perspective distortions, a non-destructive grid partitioning method was adopted. Specifically, a uniform grid of points $P = \{p_i\}_{i=1}^n$ was generated on the image to be registered, and these points were projected onto the target image using $H$ to form a corresponding point set $P' = \{p'_i\}_{i=1}^n$, resulting in a roughly aligned set of sub-images.

*2) Fine Annotation Stage.* A template matching algorithm was iteratively applied to refine the homography matrix to a precision of 1 pixel. Subsequently, all results were validated using a-contrario method [55] and manually rechecked. For any failed or suboptimal results, the annotation team manually supplemented the ground truth by marking additional points and solving for an accurate homography matrix.

*C. Data Analysis*

Compared with existing MRSI matching datasets, MapData is the largest in scale and contains the greatest number of instances, characterized by the following features:

• Diverse Types. MapData includes areas with weak texture, such as mountainous regions and water bodies, which significantly increase the matching difficulty. This diversity necessitates algorithms capable of overcoming modality differences and handling weak texture features.

• Global Coverage. MapData spans a global range with samples from various continents. The extensive geographic distribution introduces a wide variety of scene types and landscapes, thereby enhancing the generalization capability of the dataset across diverse scenarios.

• Significance. It is well known that deep learning methods heavily depend on large and varied datasets. The proposed large-scale MapData can substantially improves the generalization and stability of deep learning algorithms for MRSI matching.

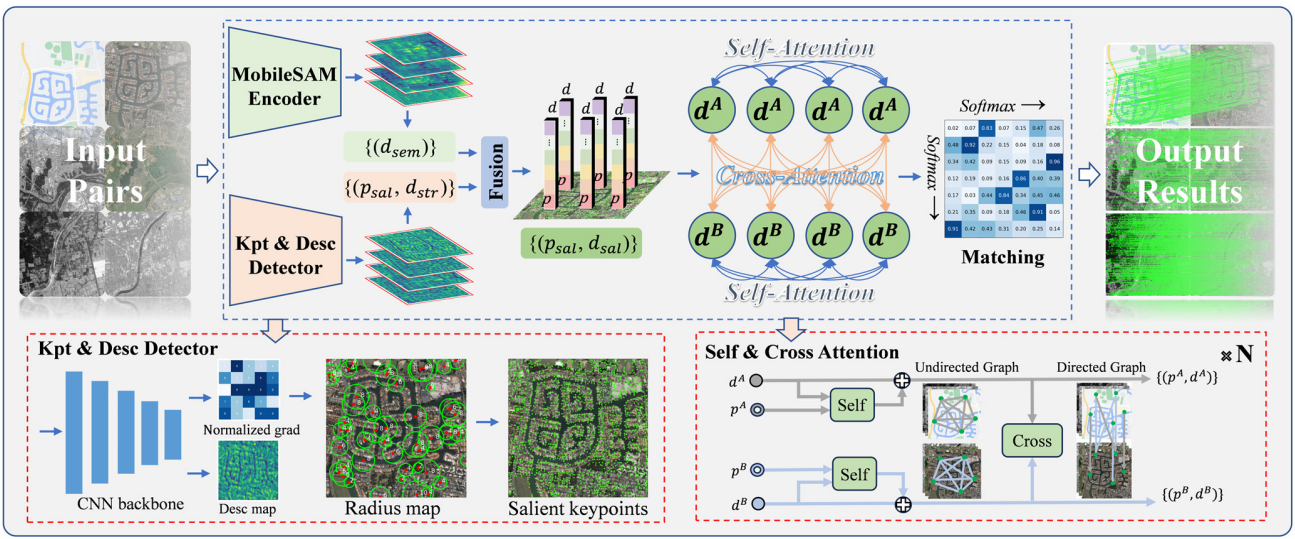

Fig. 5. Overview of the MapGlue Method. During the matching process, the pipeline begins with the feature extraction stage. In this stage, the Saliency-Enhanced SuperPoint (SES) extracts the main structural feature map and the saliency feature distribution map from the image, from which salient keypoints $p_{sal,i}$ and main structural feature descriptors $d_{str,i}$ are obtained (see Section IV.A). Next, MobileSAM [56] is used to extract detailed semantic feature descriptors $d_{sem,i}$, which are then fused with the main structural descriptors $d_{str,i}$ to produce the final saliency feature descriptors $d_{sal,i}$ (see Section IV.B). Subsequently, a series of Transformer [57] layers process these features; each layer contains several fundamental network modules that combine an intra-image undirected graph-guided self-attention mechanism with an inter-image directed graph-guided cross-attention mechanism to enhance and fuse the features (see Section IV.C). Finally, a soft matching score matrix is computed to identify corresponding keypoint pairs (see Section IV.D).

## IV. METHOD

In this paper, we propose MapGlue method, a cross-modal matching method based on a dual graph neural network that integrates semantic information. The core components of MapGlue consist of feature extraction, feature description, and feature matching, as shown in Fig. 5. A saliency constraint mechanism is introduced to regulate the distribution of keypoints, and semantic detail information is fused to enhance the semantic representation ability of feature descriptors. In addition, we constructed a dual graph structure composed of an intra-image undirected dynamic sparse graph and an inter-image directed semantic-guided graph. The former guides both global and local interactions among feature descriptors, while the latter mines potential common semantic features, significantly enhancing the robustness of matching. MapGlue effectively addresses the issues of geometric distortions and feature representation gaps in cross-modal imagery.

### A. Saliency-Enhanced SuperPoint

Compared to natural images, remote sensing images exhibit more pronounced geospatial properties and contain extensive regions with homogeneous textures and weak semantic features (e.g., rivers, lakes). Since SuperPoint [37] is trained on natural images, it does not adequately account for these characteristics, often resulting in the extraction of numerous redundant keypoints with low distinctiveness in weak-texture areas, thereby failing to effectively focus on salient regions. To address this limitation, we propose Saliency-Enhanced SuperPoint (SES) for MRSIs, building upon the SuperPoint framework. SES introduces an adaptive keypoint constraint mechanism based on saliency, which effectively guides keypoints to aggregate in regions with high semantic saliency, significantly reducing interference from redundant features in weak-texture areas and improving both keypoint distinctiveness and matching robustness.

Specifically, the SES method leverages the convolutional neural network of SuperPoint to extract the main structural feature map of an image, while simultaneously constructing a branch to extract a saliency feature distribution map. Keypoint detection and the extraction of corresponding main structural descriptors $d_{str,i}$ are performed on the main structural feature map. Meanwhile, the saliency feature distribution map adaptively guides the spatial distribution of keypoints.

*1) Saliency Feature Distribution Map Calculation.* First, SES constructs a saliency feature distribution map to characterize the saliency level of various image regions. Based on non-linear contrast enhancement of gradient information, the method quantifies the texture complexity and edge saliency of each region. The process is as follows: the input image is convolved with the Sobel operator in both the horizontal and vertical directions to obtain the corresponding gradient components, from which the gradient magnitude $G$ is computed. Subsequently, to enhance the distinction of anisotropic gradient magnitudes, a non-linear contrast adjustment is applied to highlight regions with prominent gradients, as described below:

$$G_{norm} = \left( \frac{G - G_{min}}{G_{max} - G_{min} + \epsilon} \right)^{\alpha}, \quad (1)$$

where $G_{min}$ and $G_{max}$ denote the minimum and maximum gradient magnitudes, respectively; $\epsilon = 1 \times 10^{-8}$ is a small constant to ensure numerical stability; and $a$ is the non-linear gain factor ($a = 4$; see Section V.E for ablation experiments).

*2) Adaptive Non-Maximum Suppression (ANMS).* To further suppress the emergence of redundant keypoints in weak-texture areas, an adaptive suppression radius map $R_{map}$ is constructed based on the normalized gradient magnitude $G_{norm}$, with its values dynamically adjusted according to the gradient strength. This mechanism adaptively adjusts the suppression radius for different regions, as follows:

$$R_{map} = r_{min} + (1 - G_{norm}) \times (r_{max} - r_{min}), \quad (2)$$

where $r_{min}$ and $r_{max}$ represent the minimum and maximum limits of the suppression radius (with $r_{min}$ set to the minimum pixel scale of 1 and $r_{max}$ set to 7; see Section V.E ablation experiments). In regions with dense features, a smaller suppression radius is employed to retain more keypoints; in contrast, a larger radius is used in sparse regions to reduce keypoint density.

Through these steps, an adaptive suppression radius map $R_{map}$ of the same dimensions as the input image is obtained, guiding the elimination of redundant keypoints during extraction. Consequently, the SES method not only ensures a globally uniform distribution of keypoints but also focuses the extracted keypoints on the most salient regions of the image.

*B. Fusion of Semantic Detail Information Module*

In multimodal image matching tasks, the quality of feature descriptors is critical to matching accuracy. After keypoint extraction, there is an urgent need to enhance the fusion of multimodal feature descriptors. In conventional same-modal image matching, feature descriptors are primarily constructed based on low-level visual cues (e.g., edges and corners) to build a local feature space that yields impressive matching results. However, due to significant differences in imaging mechanisms between multimodal images, relying solely on structural information fails to capture the contextual semantic information necessary for matching, which is especially evident in MRSIs, low-light images, or images with sparse textures.

Integrating both global and local semantic details is a promising approach to overcome these limitations and enhance the expressive power of feature descriptors. For example, recent methods in the image matching field, such as the SOTA method RoMa, have demonstrated exceptional stability by incorporating semantic information from DINOv2. In response, we propose a semantic-aware feature fusion method based on a dual constraint mechanism that enforces both geometric structure and semantic context. In this study, we employ a pretrained lightweight MobileSAM [56] image encoder to extract detailed semantic information with strong scene understanding capabilities. Compared to DINOv2 [43] and SAM [58], MobileSAM achieves comparable semantic understanding while maintaining a lightweight architecture. MobileSAM effectively extracts common semantic features across multimodal images, enabling the discrimination of regions with similar geometric patterns but different semantic classes (e.g., building facades versus rock textures), thus reducing the risk of mismatches.

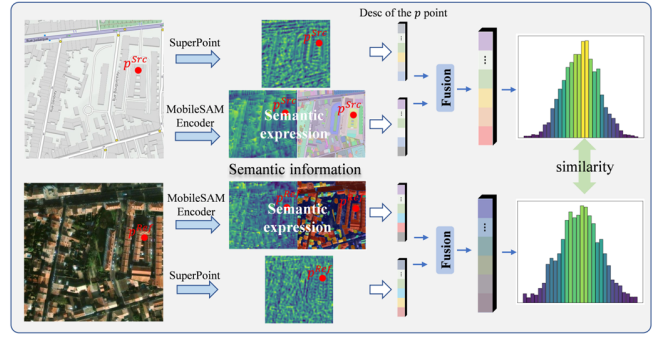

Fig. 6. Fusion of Semantic Detail Information Module. MobileSAM extracts common semantic features from multimodal images. Fusing structural and semantic descriptors yields a robust saliency descriptor.

Our method first interpolates keypoint locations on both the main structural feature map and the detailed semantic feature map to extract the main structural descriptor $d_{str,i}$ and the detailed semantic descriptor $d_{sem,i}$, respectively. These two types of descriptors are then fused to generate a saliency feature descriptor $d_{sal,i}$ that robustly represents both the semantic and structural information contained in the keypoints, as shown in Fig. 6. To fully leverage the advantages of both descriptor types, we introduce a Multi-Layer Perceptron (MLP) network for the fusion process. This fusion not only preserves the main structural information but also effectively integrates the detailed semantic information, endowing the resulting descriptor $d_{sal,i}$ with enhanced expressive power and adaptability in multimodal matching tasks. This fusion process is formulated as:

$$d_{sal,i} = \text{MLP}(Concat(d_{str,i}, d_{sem,i})), \quad (3)$$

*C. Dual Graph Structure–Guided Matching Enhancement*

In the feature matching process, extracting the global structural features of the image is essential for achieving a globally optimal matching solution. However, overly dense connection graphs often introduce redundant and valueless exchanges. This is especially problematic in MRSIs, where a large number of similar and repetitive features exist. Such redundancy not only wastes computational resources but may also interfere with matching performance. Therefore, while facilitating global information exchange, it is essential to leverage the local detailed features and semantic guidance priors both within and between images, ensuring efficient and valuable information interaction among keypoints. To this end, we propose a matching enhancement network guided by a dual graph structure by constructing an intra-image undirected dynamic sparse graph and an inter-image directed semantic-

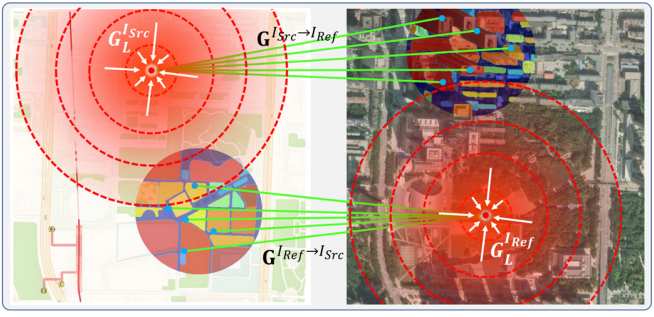

Fig. 7. Dual Graph Structure–Guided Matching Enhancement. The intra-image undirected dynamic sparse graph guides self-attention from global to local features, while the inter-image directed semantic-guided graph enhances cross-attention for common semantic feature extraction.

guided graph. Within the constructed transformer [57] layers, the intra-image graph is employed to dynamically model the spatial constraints among keypoints within a single image, facilitating effective fusion from global to local interactions. The inter-image graph is used to establish semantic associations among keypoints in different images, guiding the enhancement of cross-modal common features, as shown in Fig. 7. Each transformer layer incorporates both self-attention and cross-attention mechanisms, described as follows:

*1) Intra-Image Undirected Dynamic Sparse Graph.* Based on the Euclidean distances between keypoints, a progressive sparsification strategy is used to generate a dynamic sparse graph. A dynamic distance threshold $\epsilon$ is set, which gradually decreases as the layer index $l$ increases. This ensures that the node connections in the intra-image graph become increasingly sparse, enabling a transition from shallow global interactions to deep local interactions. Specifically, $\epsilon_l$ is computed as follows:

$$\epsilon_l = \begin{cases} \epsilon_0, & l < L/2, \\ \max\left(\epsilon_0 \cdot (1/2)^{l-L/2}, \epsilon_{\min}\right), & l \geq L/2 \end{cases} \quad (4)$$

where $\epsilon_0 = max_{i,j} d(p_i, p_j)$ is the initial global connection threshold, set to the maximum distance between keypoints to ensure that shallow layers can cover global features effectively; $\epsilon_{\min}$ is the lower bound of the distance threshold (set between 50 and 100 pixels, as detailed in Section V.E ablation experiments); $L$ is the total number of transformer layers; and $l$ is the current layer index. This design allows shallow layers ($l < L/2$) to maintain global feature interactions while deeper layers ($l \geq L/2$) progressively concentrate on local salient regions, thereby enhancing the capture of local detailed features. For each layer $l$, an intra-image graph $\mathbf{G}_l^I$ is constructed for each image based on the Euclidean distances among its keypoints and the dynamic threshold $\epsilon_l$. The construction is formulated as follows:

$$\mathbf{G}_l^I = \left(\mathbf{V}^I, \left\{(p_i^I, p_j^I) \mid d(p_i^I, p_j^I) \leq \epsilon_l\right\}\right), \quad (5)$$

where $I \in \{I_{Src}, I_{Ref}\}$, $\mathbf{G}_l^{I_{Src}}$ and $\mathbf{G}_l^{I_{Ref}}$ represent the intra-image graphs for the source image and the reference image, respectively; $\mathbf{V}^I$ denote the sets of all keypoints in these images; and for any two keypoints $p_i^I$ and $p_j^I$, the Euclidean distance $d(p_i^I, p_j^I)$ is computed. During attention interactions, the adjacency relationships defined by $\mathbf{G}_l^I$ are used to compute attention scores between nodes. Since the intra-image graph is undirected, attention can be exchanged in both directions between any two keypoints. The attention score is calculated as follows:

$$\alpha_{ij} = \langle q_i, \phi(p_i, p_j) k_j \rangle, \quad (6)$$

where $p_i$ and $p_j$ denote the positions of the keypoints, the operator $\langle \cdot, \cdot \rangle$ denotes the inner product, and $\phi(\cdot)$ represents a rotational position encoding function. We introduce rotational position encoding to better accommodate rotation invariance. The rotational position encoding is defined as:

$$\phi(p_i, p_j) = \text{Diag}\left(\psi(b_1^\top(p_j - p_i)), \psi(b_2^\top(p_j - p_i)), \ldots, \psi(b_{d/2}^\top(p_j - p_i))\right) \quad (7)$$

where $\{b_k\}$ is a set of predefined orthogonal basis vectors, and $\psi(\cdot)$ maps a scalar into a rotational embedding matrix. This encoding enables the attention mechanism to capture the relative positional relationships between keypoints, maintaining invariance under image rotation and translation.

For $I \in \{I_{Src}, I_{Ref}\}$, each keypoint $i$ aggregates information from its neighboring nodes according to the self-attention scores to compute the aggregated feature $m_i$ as follows:

$$m_i = \frac{\sum_{j \in G(i)} \exp(a_{ij}) W_x x_j}{\sum_{j \in G(i)} \exp(a_{ij})}, \quad (8)$$

where $G(i)$ is the set of neighboring keypoints of $i$ in the intra-image graph, $W_x$ is the weight matrix used for linear transformation, and $x_j$ denotes the feature vector of keypoint $j$.

*2) Inter-Image Directed Semantic-Guided Graph.* In cross-image attention interactions, the semantic similarity is computed based on the detailed semantic descriptors $d_{sem,i}$ of keypoints. Each keypoint in one image computes its semantic similarity with every keypoint in the other image to construct a directed graph between the two images. Since the inter-image graph is directional, with information flowing from source nodes to target nodes, two separate directed inter-image graphs, $\mathbf{G}^{I_{Src} \to I_{Ref}}$ and $\mathbf{G}^{I_{Ref} \to I_{Src}}$, are constructed separately. By establishing adjacency relationships among nodes with high semantic similarity, common cross-modal features can be identified. Taking the construction of $\mathbf{G}^{I_{Src} \to I_{Ref}}$ as an example, the inter-image graph is defined as:

$$\mathbf{G}^{I_{Src}, I_{Ref}} = \left(\mathbf{V}^{I_{Src}} \cup \mathbf{V}^{I_{Ref}}, \left\{\left(p_i^{I_{Src}}, p_j^{I_{Ref}}\right) \mid sim\left(d_{sem,i}^{I_{Src}}, d_{sem,j}^{I_{Ref}}\right) \in TopK\right\}\right) \quad (9)$$

where *TopK* indicates that an adjacency relationship is established between the source node and the candidate points with the top 50% semantic similarity; and $sim(\cdot, \cdot)$ denotes the cosine similarity function between semantic descriptors. The construction of $\mathbf{G}^{I_{Ref} \to I_{Src}}$ is analogous to that of $\mathbf{G}^{I_{Src} \to I_{Ref}}$.

Under the semantic constraints of $\mathbf{G}^{I_{Src}, I_{Ref}}$, during the cross-attention computation between the two images, each source node interacts only with those keypoints in the other image that exhibit high semantic similarity. This selective cross-attention minimizes the learning of redundant interference from multimodal images and aims to extract the common features between the two images to achieve optimal matching.

Finally, after each transformer layer completes its self-attention and cross-attention interactions, a residual connection is incorporated for feature updating. This is computed as follows:

$$x_i \leftarrow x_i + \text{MLP}([\text{Concat}(x_i, m_i)]), \qquad (10)$$

This additional shortcut ensures effective feature fusion while preventing degradation during deeper network training.

*D. Constructing the Soft Matching Score Matrix*

After the robust feature vectors $\mathbf{X}^{Src}$ and $\mathbf{X}^{Ref}$ have been obtained through the dual graph structure guided self-attention and cross-attention mechanisms described in Section IV.C, a soft matching score matrix is constructed to compute the final matching results. First, $\mathbf{X}^{Src}$ and $\mathbf{X}^{Ref}$ are individually subjected to a linear transformation, and then the matching score matrix $\mathbf{S}$ between all pairs of keypoints in the two images is computed as defined in Equation (11):

$$\mathbf{S} = \langle Linear(\mathbf{X}^{Src}), Linear(\mathbf{X}^{Ref}) \rangle, \qquad (11)$$

where $Linear(\cdot)$ is a linear mapping operation, and $\langle \cdot, \cdot \rangle$ represents the inner product, which is used to measure the similarity between two feature vectors.

Subsequently, to ensure the reliability and stability of feature matching, a dual-softmax strategy is employed. Specifically, the softmax function is applied along both the row and column dimensions of matrix $\mathbf{S}$ to obtain the normalized soft matching score matrix $\mathbf{P}$, as computed in Equation (12):

$$\mathbf{P} = \text{Softmax}_{row}(\mathbf{S}) \cdot \text{Softmax}_{col}(\mathbf{S}), \qquad (12)$$

This bidirectional normalization ensures consistency and uniqueness in matching from the source image $I_{Src}$ to the reference image $I_{Ref}$ and vice versa, so that each pair of keypoints receives an optimal matching score mutually validated in both directions. Finally, by setting a threshold $\tau$ [41], the soft matching score matrix $\mathbf{P}$ is compared against $\tau$ to filter high-quality feature matches, and a mutual nearest neighbor (MNN) criterion is used to establish one-to-one correspondences between keypoints.

*E. Supervision*

We use the ground-truth homography matrices from the MapData dataset as training labels. Specifically, a soft matching ground-truth score matrix is constructed between keypoints using the ground-truth homography $H$. Since the imaging mechanisms of electronic navigation maps and visible light images differ, their pixels cannot be perfectly aligned on a one-to-one basis. Therefore, we adopt a bidirectional reprojection supervision mechanism. In this mechanism, keypoints extracted from both images are projected bidirectionally using $H$. We set a positive sample threshold $th_{pos} = 3$ and a negative sample threshold $th_{neg} = 6$; only keypoints whose bidirectional projection distances are less than $th_{pos}$ are considered positive samples, while those with distances greater than $th_{neg}$ are treated as negative samples. Keypoints with distances between these thresholds are considered ambiguous and are ignored during loss computation. This strategy is designed to improve the model's generalization and better uncover potential feature matching relationships.

To effectively optimize the training process of the feature matching model, we design a cross-entropy loss function based on multiple supervisory signals. This loss function not only considers the matching performance for positive and negative samples but also introduces penalties for false positives (FP) and false negatives (FN) through a quadruplet supervision strategy, thereby simultaneously optimizing matching accuracy and robustness. The design goal of the loss function is to maximize the confidence of positive samples while suppressing the responses of negative samples and to impose gradient penalties on erroneous matches, ultimately enhancing the model's discriminative ability and generalization performance. The overall loss consists of four components: positive sample loss $L_{pos}$, negative sample loss $L_{neg}$, false positive loss $L_{fp}$, and false negative loss $L_{fn}$, as shown in Equation (13):

$$L = L_{pos} + \frac{1}{3}\left(L_{neg} + L_{fp} + L_{fn}\right), \qquad (13)$$

where the individual loss terms are defined in Equation (14) as follows:

$$\begin{cases} L_{pos} = -\dfrac{1}{N_{pos}} \sum\limits_{i \in P} \log(p_i) \\ L_{neg} = -\dfrac{1}{N_{neg}} \sum\limits_{j \in N} \log(1 - p_j) \\ L_{fp} = -\dfrac{1}{N_{fp}} \sum\limits_{k \in FP} \log(1 - p_k) \\ L_{fn} = -\dfrac{1}{N_{fn}} \sum\limits_{m \in FN} \log(p_m) \end{cases}, \qquad (14)$$

where $p$ denotes the set of positive samples; $N$ represents the set of negative samples; $FP$ denotes the set of false positive samples; $FN$ denotes the set of false negative samples; and $p_i$, $p_j$, $p_k$, $p_m$ represent the predicted confidence scores for the corresponding samples.

## V. EXPERIMENTS

### A. Implementation Details

Training was conducted on an NVIDIA RTX A6000 GPU, while evaluation was performed on an RTX 4090D GPU. During training, we used the Adam optimizer with an initial learning rate of 1e-4 and a batch size of 16. The network was trained to convergence over approximately 80 hours. To ensure consistency across experiments, all training images were resized to 512×512 pixels. Hyperparameters include the non-linear gain factor $a$, the maximum suppression radius $r_{max}$, and the minimum distance threshold $\epsilon_{min}$ (detailed in Section V.E ablation experiments). The MapGlue model was trained solely on the MapData dataset and then evaluated for generalization performance on other multimodal datasets. During the training phase, random perspective transformations were applied as data augmentation, with specific parameters: rotation angles in the range (-180°, 180°), scale variations in the range (0.5, 1.5), and translations in the range (0, 0.5).

*1) Evaluation Datasets* To comprehensively and objectively assess the proposed MapGlue method, experiments were conducted on multiple datasets covering various scenes, categories, and modalities. The evaluation datasets are detailed as follows:

• MapData-test. This dataset comprises 1910 pairs of map–visible image pairs, covering a variety of typical geographic scenes including urban areas, mountainous regions, water bodies, and deserts. The matching ground truth is provided in the form of homography matrices.

• MRSI Matching Datasets. The SRIF dataset contains 6 different modalities, each with 200 image pairs (totaling 1200 pairs), with matching ground truth provided as affine transformation matrices. The CoFSM dataset also comprises 6 different modalities datasets, each containing 10 image pairs (totaling 60 pairs), with matching ground truth provided as real corresponding keypoint pairs. The MultiResSAR-Low dataset consists of 1000 pairs of sar–visible image pairs, with matching ground truth provided as homography matrices.

• Multimodal Natural Image (MNI) Matching Datasets. The LGHD LWIR/RGB dataset contains 44 pairs of aligned visible and thermal infrared building images. The DIODE test dataset comprises 612 pairs of aligned depth-visible images of indoor and outdoor buildings.

*2) Evaluation Metrics.* The evaluation datasets contain two types of matching labels: (1) labels based on homography matrices and (2) labels where the images are pre-aligned. For scenes where homography matrices serve as labels, the four corner points of the source image are projected onto the target image using both the ground-truth and predicted homography matrices, and the average projection error is computed. For cases where the images are already aligned, a simulated homography transformation is applied to the source image, and the predicted homography is used to restore the alignment. In this process, RANSAC is used as a robust homography estimator, with a threshold of 1.5, 10K iterations, and a confidence level of 0.9999. Subsequently, the Area Under the Curve (AUC) is computed for different pixel error thresholds (3px, 5px, and 10px) to quantitatively evaluate the accuracy of the homography predictions. To ensure fairness across different methods, if the longer side of a test image exceeds 640 pixels, it is uniformly scaled down to 640 pixels for evaluation. Furthermore, to comprehensively assess the generalization performance of each method under varying matching difficulties, three levels of simulated transformations (Easy, Normal, Hard) are applied to the source images in each dataset. For each transformation type, experiments are repeated 5 times with random variations, and the average performance is reported. The specific transformation rules are as follows:

• **Easy Transformation.** Up to 10% random transformation is applied, including rotations within (-36°, 36°), translations along the x and y axes (within ±10% of the image width and height, respectively), and scale variations in the range (0.9, 1.1).

• **Normal Transformation.** Up to 20% random transformation is applied, including rotations within (-72°, 72°), translations along the x and y axes (within ±20% of the image width and height, respectively), and scale variations in the range (0.8, 1.2).

• **Hard Transformation.** To ensure that sufficient common regions exist between image pairs, a maximum of 50% random rotation is applied, with rotation angles in the range (-180°, 180°). The rotation is applied without altering the size of the common regions. In addition, translations and scale variations are limited to within 30% (i.e., translations along the x and y axes are within ±30% of the image width and height, and scale variations are in the range (0.7, 1.3)).

*3) Baseline Methods.* Current mainstream feature matching algorithms can be categorized into sparse matching, semi-dense matching, and dense matching. For sparse matching methods, we compared three advanced traditional multimodal matching methods (RIFT, HOWP, and POS-GIFT) as well as representative deep learning methods (D2-Net, SuperGlue, LightGlue, $GIM_{LG}$, and $MINIMA_{LG}$). To ensure fairness, the number of keypoints extracted by all sparse matching methods is uniformly set to 2048. For semi-dense matching methods, we selected LoFTR, ELoFTR, $GIM_{LoFTR}$, $MINIMA_{LoFTR}$, and XoFTR (which is capable of cross-modal image matching). For dense matching methods, we compared DKM, $GIM_{DKM}$, RoMa, and $MINIMA_{RoMa}$. Here, $MINIMA_{LG}$, $MINIMA_{LoFTR}$, and $MINIMA_{RoMa}$ are versions of LightGlue, LoFTR, and RoMa retrained on multimodal datasets. For the semi-dense and dense matching methods, the number of keypoints is left unchanged (using the default parameters of each model). We compare the aforementioned methods with the proposed MapGlue as well as a lighter version without MobileSAM semantic information, termed FastMapGlue.

### B. Evaluate on Our MapData

As shown in Table II, all methods exhibit a significant performance degradation trend from Easy to Normal to Hard transformations. Although the traditional methods RIFT, HOWP, and POS-GIFT perform poorly overall, their performance is comparatively more robust than that of deep learning methods trained solely on visible light images (e.g., D2-Net, SuperGlue, LightGlue, and GIM$_{LG}$). This indicates that deep learning algorithms trained exclusively on visible light imagery suffer a substantial drop in matching performance when generalized to scenarios with large modality differences such as map-visible images. On the other hand, MINIMA$_{LG}$, which is trained on multimodal data, demonstrates a marked improvement under the Easy transformation, highlighting the advantages of multimodal data-driven approaches. However, its AUC@5px drops by 87.8% from Easy to Hard, reaching only 3.13% under Hard conditions indicating a lack of stability in complex scenarios.

TABLE II. Homography Evaluation on the MapData-test Dataset under Three Levels of Transformation Difficulty. The AUC of the projective error is reported as a percentage.

| Category | Method | Easy | | | Normal | | | Hard | | |
|---|---|---|---|---|---|---|---|---|---|---|
| | | @3px | @5px | @10px | @3px | @5px | @10px | @3px | @5px | @10px |
| Sparse | RIFT [28] (TIP 2019) | 0.24 | 0.75 | 3.46 | 0.07 | 0.16 | 0.98 | 0 | 0.06 | 0.31 |
| | HOWP [32] (ISPRS 23) | 2.2 | 7.83 | 21.66 | 0.35 | 2.02 | 6.66 | 0 | 0.23 | 1.23 |
| | POS-GIFT [35] (Inf. Fusion 23) | 1 | 4.5 | 15.82 | 0.29 | 2.23 | 10.37 | 0.11 | 0.85 | 5.45 |
| | D2-Net [39] (CVPR 19) | 1.71 | 5.73 | 13.37 | 0.39 | 1.61 | 4.7 | 0.1 | 0.41 | 1.27 |
| | SuperGlue [40] (CVPR 20) | 0.65 | 2.27 | 5.54 | 0.22 | 0.83 | 2.39 | 0.06 | 0.2 | 0.57 |
| | LightGlue [41] (ICCV 23) | 0.16 | 0.26 | 0.62 | 0 | 0.04 | 0.17 | 0.03 | 0.08 | 0.17 |
| | GIM$_{LG}$ [53] (ICLR 24) | 0.03 | 0.16 | 0.67 | 0.01 | 0.07 | 0.39 | 0.02 | 0.06 | 0.24 |
| | MINIMA$_{LG}$ [54] (CVPR 2025) | 13.84 | 25.59 | 39.91 | 5.56 | 11.77 | 20.14 | 1.42 | 3.13 | 5.89 |
| Semi-Dense | LoFTR [46] (CVPR 21) | 0.14 | 0.95 | 4.89 | 0.04 | 0.27 | 1.54 | 0.05 | 0.13 | 0.46 |
| | ELoFTR [47] (CVPR 24) | 0.11 | 0.42 | 1.73 | 0.03 | 0.14 | 0.67 | 0 | 0.03 | 0.15 |
| | XoFTR [18] (CVPR 24) | 10.00 | 20.79 | 35.79 | 3.9 | 9.2 | 17.65 | 0.94 | 2.39 | 5.12 |
| | GIM$_{LoFTR}$ [53] (ICLR 24) | 0 | 0.02 | 0.05 | 0 | 0 | 0.01 | 0.05 | 0.13 | 0.46 |
| | MINIMA$_{LoFTR}$ [54] (CVPR 2025) | 0.66 | 2.04 | 5.11 | 0.19 | 0.55 | 1.7 | 0.06 | 0.2 | 0.52 |
| Dense | DKM [49] (CVPR 23) | 0.02 | 0.06 | 0.27 | 0 | 0.02 | 0.12 | 0 | 0 | 0.03 |
| | GIM$_{DKM}$ [53] (ICLR 24) | 0 | 0.03 | 0.27 | 0 | 0.02 | 0.13 | 0 | 0.01 | 0.07 |
| | RoMa [50] (CVPR 24) | 0.88 | 2.49 | 6.35 | 0.4 | 1.17 | 2.99 | 0.13 | 0.37 | 0.95 |
| | MINIMA$_{RoMa}$ [54] (CVPR 2025) | 14.34 | 24.96 | 38.48 | 5.67 | 10.78 | 18.24 | 1.66 | 3.26 | 6.07 |
| Ours | FastMapGlue | **36.75** | **54.65** | **73.27** | **26.68** | **43.81** | 63.74 | 14.57 | 27.64 | 46.25 |
| | MapGlue | 33.08 | 51.71 | 71.93 | 25.23 | 42.71 | **64.09** | **19.03** | **34.6** | **54.99** |

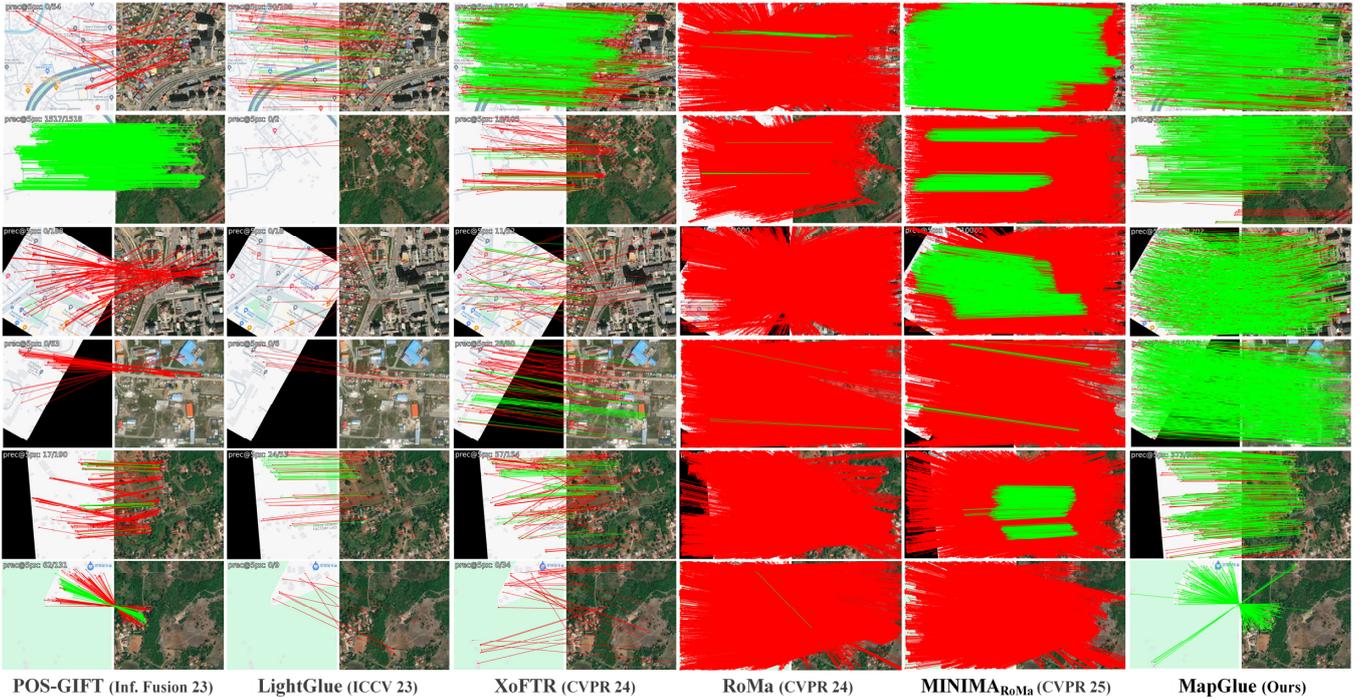

Fig. 8. Qualitative Results on Map–Visible Image Pairs (red lines indicate matches with a projective error greater than 5 pixels).

Among the semi-dense matching methods, models such as LoFTR, ELoFTR, and GIM$_{LoFTR}$ tend to overfit to visible light images and are unable to generalize to the significant domain shift in map–visible image pairs, resulting in AUC@5px values below 1% across all transformation difficulties. XoFTR, benefiting from extensive training on infrared–visible image pairs and an advanced network design, performs reasonably well on the MapData-test under Easy transformation. However, its AUC@5px drops by 88.5% from Easy to Hard, showing considerable impact in complex scenarios. MINIMA$_{LoFTR}$, despite being trained on multimodal data, still faces clear domain shift issues on the real-world MapData-test dataset and, as an earlier classic network, its performance remains unsatisfactory.

Table III. Homography Evaluation on MultiResSAR-Low, SRIF, and CoFSM Datasets under Three Transformation Difficulties. The AUC for the projective error (AUC@5px) is reported as a percentage. Only the AUC@5px values are shown.

| Category | Method | MultiResSAR-Low | | | SRIF | | | CoFSM | | |
|---|---|---|---|---|---|---|---|---|---|---|
| | | Easy | Normal | Hard | Easy | Normal | Hard | Easy | Normal | Hard |
| Sparse | RIFT [28](TIP 2019) | 1.22 | 0.41 | 0.1 | 0.63 | 0.21 | 0.12 | 3.48 | 1.53 | 0.52 |
| | HOWP [32](ISPRS 23) | 3.6 | 0.87 | 0.1 | 2.68 | 1.07 | 0.32 | 15.27 | 5.84 | 0.86 |
| | POS-GIFT [35](Inf. Fusion 23) | 3.02 | 1.77 | 0.81 | 3.42 | 2.27 | 1.56 | 14.38 | 10.38 | 4.87 |
| | D2-Net [39](CVPR 19) | 2.41 | 0.83 | 0.15 | 1.28 | 0.77 | 0.38 | 12.45 | 5.09 | 1.29 |
| | SuperGlue [40](CVPR 20) | 6.7 | 3.27 | 0.91 | 5.4 | 3.89 | 1.57 | 29.35 | 17.64 | 5.4 |
| | LightGlue [41](ICCV 23) | 5.49 | 2.41 | 0.56 | 4.26 | 3.22 | 1.24 | 29.18 | 12.75 | 3.26 |
| | GIM$_{LG}$ [53](ICLR 24) | 8.35 | 6.18 | 2.47 | 5.35 | 4.54 | 3.18 | 28.64 | 22.56 | 12.53 |
| | MINIMA$_{LG}$ [54](CVPR 2025) | 13.68 | 5.74 | 1.62 | 12.53 | 8.49 | 3.22 | 35.41 | 16.9 | 5.86 |
| Semi-Dense | LoFTR [46](CVPR 21) | 1.79 | 0.75 | 0.33 | 1.57 | 0.92 | 0.73 | 20.44 | 7.57 | 4.54 |
| | ELoFTR [47](CVPR 24) | 2.71 | 1.45 | 0.35 | 2.17 | 1.33 | 0.39 | 9.52 | 2.68 | 0.98 |
| | XoFTR [18](CVPR 24) | 7.62 | 3.12 | 1.01 | 6.13 | 3.61 | 1.35 | 29.13 | 14.29 | 3.89 |
| | GIM$_{LoFTR}$ [53](ICLR 24) | 1.29 | 0.67 | 0.27 | 1.41 | 1.09 | 1.69 | 15.88 | 9.47 | 4.3 |
| | MINIMA$_{LoFTR}$ [54](CVPR 2025) | 3.41 | 1.06 | 0.29 | 1.95 | 0.92 | 0.32 | 15.03 | 8.26 | 1.1 |
| Dense | DKM [49](CVPR 23) | 3.76 | 1.54 | 0.5 | 5.45 | 4.3 | 1.83 | 23.69 | 12.35 | 5.26 |
| | GIM$_{DKM}$ [53](ICLR 24) | 3.79 | 2.82 | 1.66 | 6.21 | 5.95 | 4.68 | 29.88 | 23.32 | 17.65 |
| | RoMa [50](CVPR 24) | 12.57 | 5.66 | 1.65 | 10.69 | 7.47 | 3.39 | 33.57 | 24.38 | 7.57 |
| | MINIMA$_{RoMa}$ [54](CVPR 2025) | **23.8** | 12.06 | 3.45 | 17.22 | 11.99 | 5.05 | **42.82** | 27.42 | 13.6 |
| Ours | FastMapGlue | 20.36 | 14.38 | 9.66 | 23.57 | 18.97 | 9.84 | 36.22 | 30.32 | 20.59 |
| | MapGlue | 23.29 | **17.52** | **12.82** | **24.23** | **19.54** | **13.61** | 38.01 | **32.29** | **25.58** |

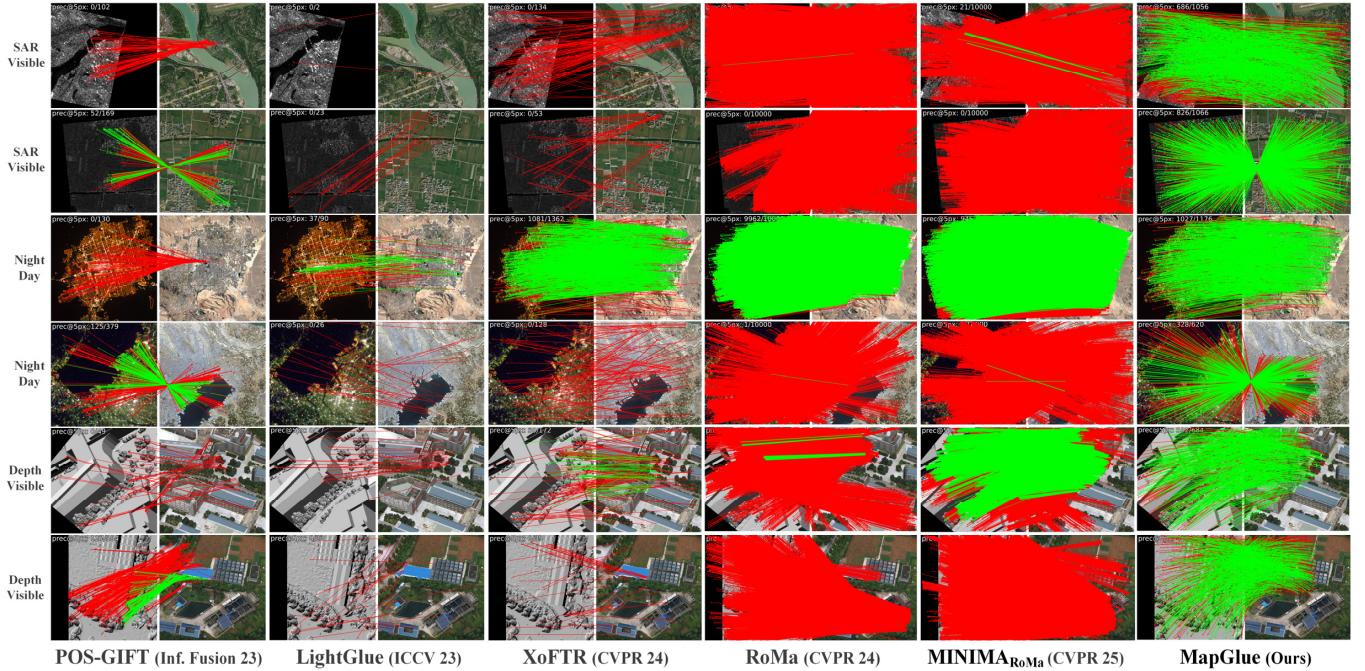

Fig. 9. Qualitative Results of Multimodal Remote Sensing Image Matching (red lines indicate matches with a projective error greater than 5 pixels).

In the dense matching category, both DKM and GIMDKM perform poorly when confronted with datasets exhibiting large modality differences. The RoMa method, which integrates DINOv2 semantic information, outperforms DKM overall, while MINIMA$_{RoMa}$ trained specifically for multimodal matching achieves the best performance among the compared methods. However, even MINIMA$_{RoMa}$ experiences an 86.9% drop in AUC@5px under Hard transformation, reflecting its limited stability under complex conditions.

Comparatively, our proposed FastMapGlue significantly outperforms other methods under both Easy and Normal transformations, with its AUC@5px improving by 119% and 306.4% relative to MINIMA$_{RoMa}$, respectively. Moreover, MapGlue which integrates MobileSAM semantic information shows particularly notable performance under Hard transformations, with its AUC@5px improving by 961.3% compared to the best-performing $MINIMA_{RoMa}$. This result demonstrates that incorporating semantic information markedly enhances matching precision and stability in complex scenarios. The gain from semantic information is expected to be even more pronounced when generalized to other multimodal datasets. Overall, these results confirm that, driven by the MapData dataset and optimized network design, MapGlue effectively addresses the registration challenges induced by geometric, radiometric, scale, and rotation differences between map-visible image pairs, and exhibits strong stability and excellent performance in multimodal matching tasks.

TABLE IV. Homography Evaluation on the DIODE and LGHD LWIR/RGB Datasets under Three Transformation Difficulties. The AUC for the projective error (AUC@5px) is reported as a percentage, with only the AUC@5px values (i.e., within 5 pixels of error) shown.

| Category | Method | DIODE | | | LGHD LWIR/RGB | | |
|---|---|---|---|---|---|---|---|
| | | Easy | Normal | Hard | Easy | Normal | Hard |
| Sparse | RIFT [28](TIP 2019) | 1.28 | 0.38 | 0.12 | 2.54 | 0.68 | 1.15 |
| | HOWP [32](ISPRS 23) | 4.02 | 10.7 | 0.23 | 16.02 | 4.29 | 2.62 |
| | POS-GIFT [35](Inf. Fusion 23) | 2.97 | 1.69 | 0.69 | 9.82 | 6.28 | 2.47 |
| | D2-Net [39](CVPR 19) | 3.62 | 1.16 | 0.32 | 5.3 | 2.25 | 0.85 |
| | SuperGlue [40](CVPR 20) | 11.21 | 4.25 | 1.24 | 19.16 | 9.9 | 2.63 |
| | LightGlue [41](ICCV 23) | 6.06 | 2.45 | 0.58 | 12.63 | 3.74 | 1.86 |
| | $GIM_{LG}$ [53](ICLR 24) | 8.4 | 5.58 | 2.48 | 13.57 | 6.45 | 3.39 |
| | $MINIMA_{LG}$ [54](CVPR 2025) | 34.31 | 9.17 | 4.48 | 28.13 | 13.28 | 4.11 |
| Semi-Dense | LoFTR [46](CVPR 21) | 6.82 | 2.22 | 0.83 | 5.25 | 2.3 | 0.68 |
| | ELoFTR [47](CVPR 24) | 9.06 | 3.34 | 0.83 | 3.51 | 0.96 | 0.38 |
| | XoFTR [18](CVPR 24) | 32.55 | 13.97 | 3.91 | 27.31 | 11.21 | 2.66 |
| | $GIM_{LoFTR}$ [53](ICLR 24) | 0.12 | 0.11 | 0 | 1.18 | 0.64 | 0.25 |
| | $MINIMA_{LoFTR}$ [54](CVPR 2025) | 8.17 | 3.19 | 1 | 8.27 | 4.46 | 0.63 |
| Dense | DKM [49](CVPR 23) | 2.95 | 1.64 | 0.32 | 6.95 | 3.3 | 0.79 |
| | $GIM_{DKM}$ [53](ICLR 24) | 4.05 | 2.35 | 1.01 | 11.1 | 8.68 | 4.36 |
| | RoMa [50](CVPR 24) | 27.18 | 13.96 | 4.68 | 22.63 | 12.04 | 3.11 |
| | $MINIMA_{RoMa}$ [54](CVPR 2025) | **43.03** | **25.85** | 8.46 | 28.24 | 16.88 | 5.45 |
| Ours | FastMapGlue | 29.28 | 19.34 | 9.71 | **29.04** | 18.91 | 13.61 |
| | MapGlue | 33.83 | 25.3 | **15.83** | 28.45 | **20.17** | **15.57** |

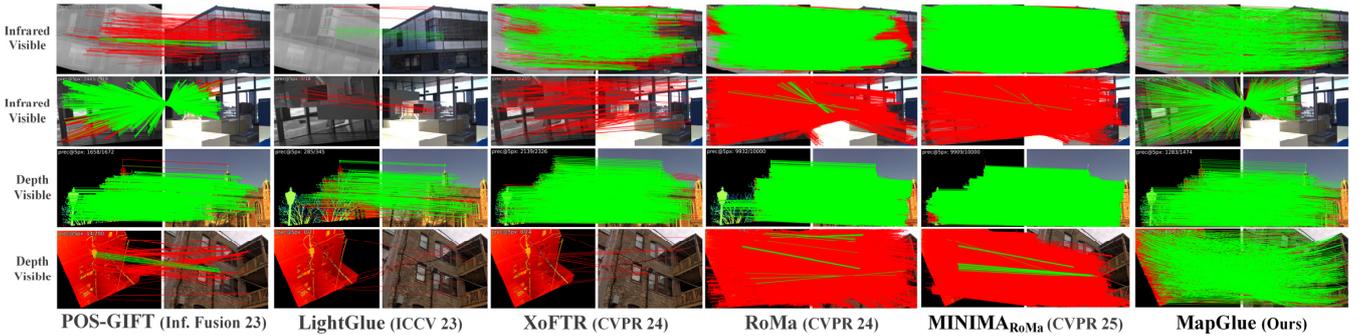

Fig. 10. Qualitative Results on Multimodal Natural Image Matching (red lines indicate matches with a projective error greater than 5 pixels).

In qualitative experiments on the MapData-test dataset, representative and SOTA algorithms were selected from each category—traditional sparse matching (POS-GIFT), deep learning sparse matching (LightGlue), semi-dense matching (XoFTR), and dense matching (RoMa and $MINIMA_{RoMa}$) and compared with the proposed MapGlue. As shown in Fig. 8, due to the significant modality gap in map–visible image pairs, the matching difficulty is greatly increased, and the competing algorithms struggle to maintain good stability under such conditions. In contrast, MapGlue consistently achieves high-precision and stable matching results in both simple and complex scenarios.

### C. Evaluate on MRSI

In addition, we conducted generalization tests on other MRSI datasets. It is noteworthy that our method was trained solely on the map–visible modality dataset. As shown in Table III, under the Easy transformation, the multimodal model $MINIMA_{RoMa}$ achieves the best performance on the MultiResSAR-Low and CoFSM datasets, while under the Normal and Hard transformations, MapGlue demonstrates significantly stronger generalization capability. On the MultiResSAR-Low dataset, MapGlue improves AUC@5px by at least 45.3% and 271.6% under Normal and Hard transformations, respectively. On the SRIF dataset, MapGlue's AUC@5px increases by 40.7%, 63%, and 169.5% under the Easy, Normal, and Hard transformations, respectively. In the CoFSM dataset, MapGlue's AUC@5px improves by 17.8% and 88.1% under Normal and Hard transformations, respectively. These results clearly demonstrate that MapGlue not only excels on the map–visible modality but also exhibits excellent generalization capability on other modalities.

TABLE V Ablation study of MapGlue.

| Method | MapData-test-Hard | | | MultiResSAR-Low-Easy | | |
|---|---|---|---|---|---|---|
| | @3px | @5px | @10px | @3px | @5px | @10px |
| (1) FastMapGlue with Superpoint | 14.19 | 26.48 | 43.91 | 4.99 | 16.04 | 36.69 |
| (2) FastMapGlue with SES | 14.57 | 27.64 | 46.25 | 6.5 | 20.36 | 43.65 |
| (3) MapGlue without Graph with Superpoint | 15.25 | 28.77 | 47.66 | 5.9 | 17.45 | 38.65 |
| (4) MapGlue with Superpoint | 16.27 | 30.33 | 50.02 | 6.56 | 18.6 | 39.59 |
| (5) MapGlue without Graph | 16.36 | 30.68 | 49.8 | 7.78 | 21.82 | 44.02 |
| (6) Full (MapGlue) | **19.03** | **34.6** | **54.99** | **8.66** | **23.29** | **45.99** |

TABLE VI. Quantitative Experimental Results Comparing SuperPoint and SES

| Method | MapData-test-Hard | | | MultiResSAR-Low-Easy | | |
|---|---|---|---|---|---|---|
| | @3px | @5px | @10px | @3px | @5px | @10px |
| FastMapGlue with Superpoint | 219 | 348 | 465 | 136 | 250 | 398 |
| FastMapGlue with SES | **338** | **569** | **739** | **304** | **583** | **892** |
| MapGlue with Superpoint | 254 | 397 | 521 | 141 | 254 | 392 |
| MapGlue with SES | **388** | **645** | **830** | **323** | **608** | **910** |

Qualitative comparisons between MapGlue and other representative methods, including POS-GIFT, LightGlue, XoFTR, RoMa, and MINIMA$_{RoMa}$, on multimodal remote sensing images are shown in Fig. 9. Despite being trained solely on the map–visible modality dataset, MapGlue consistently achieves high-precision and robust matching performance across different multimodal remote sensing scenarios. Essentially, MapGlue accurately captures cross-modal common features in complex scenes, effectively overcoming various modality differences, whereas the competing methods exhibit significant mismatches or even complete failure in certain cases. This demonstrates the outstanding generalization ability and high stability of MapGlue in cross-modal matching tasks.

*D. Evaluate on MNI*

To further validate the generalization ability of MapGlue, we conducted tests not only on remote sensing datasets but also on MNI datasets. Since MapGlue was not trained on natural images, its performance on natural imagery shows a slight decrease. As shown in Table IV, in the DIODE dataset, the MINIMA$_{RoMa}$ model achieves the best performance under Easy and Normal transformations; however, under the Hard transformation where the scenes are more complex, MapGlue performs superiorly, with its AUC@5px improving by 87.1%. In the LGHD LWIR/RGB dataset, MapGlue outperforms other models in AUC@5px by 2.8%, 19.5%, and 185.7% under Easy, Normal, and Hard transformations, respectively, with the performance gain under Hard transformations being particularly striking. Although methods such as MINIMA$_{LG}$, MINIMA$_{RoMa}$, and XoFTR perform well under Easy transformations, their performance deteriorates significantly under Normal and Hard conditions. In contrast, the proposed FastMapGlue and MapGlue methods demonstrate much stronger stability and robustness, with the semantic-enhanced MapGlue exhibiting superior stability in handling complex architectural scenes.

As shown in Fig. 10, while all algorithms perform well on image pairs with low matching difficulty, most competing methods tend to fail on high-difficulty pairs, whereas MapGlue maintains both stability and accuracy.

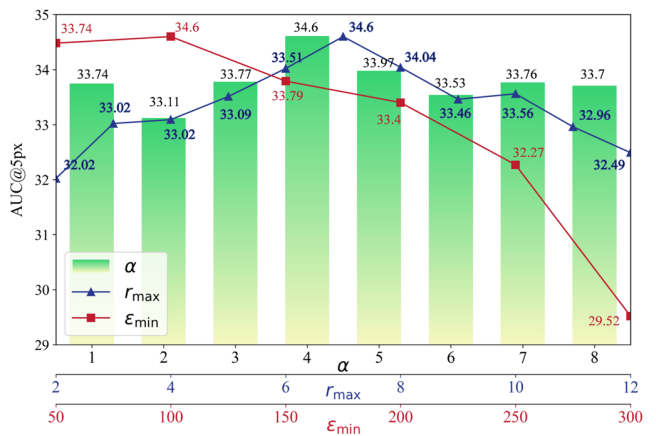

Fig. 11. Ablation Experiments of Hyperparameters $a$, $r_{max}$ and $\epsilon_{min}$. The AUC for the projective error (AUC@5px) is reported as a percentage, with only the AUC@5px values shown.

*E. Ablation Studies*

To validate the rationale behind our hyperparameter settings, we evaluated three key hyperparameters on the challenging MapData-test-Hard dataset: the non-linear gain factor $a$, the maximum suppression radius $r_{max}$, and the minimum distance threshold $\epsilon_{min}$. As shown in Fig. 11, quantitative experimental results indicate that during the keypoint extraction process, applying non-linear contrast gradient adjustment with $a$ set to 4 yields the best performance; during adaptive non-maximum suppression, a maximum suppression radius $r_{max}$ of 7 is optimal; and in the feature interaction module, constructing the intra-image undirected dynamic sparse graph with a minimum local interaction threshold $\epsilon_{min}$ in the range of 50 to 100 pixels produces the best overall results.

To systematically validate the contributions of the modules proposed in this paper, we evaluated six different variants of MapGlue on the MapData-test-Hard and MultiResSAR-Low Easy datasets. We compared the performance of the proposed SES module with that of SuperPoint and specifically examined the contribution of the dual graph structure to enhancing model stability (quantitative results are shown in Table V). By comparing the results of variants (2) with (1) and (6) with (4), it is evident that the SES module provides more accurate position estimation for subsequent feature matching, which significantly improves the model's matching accuracy and stability. Furthermore, comparing variant (6) with (5) further confirms the significant role of the dual graph structure in reinforcing model stability. In addition, the comparison between (6) and (2) demonstrates that, after fusing MobileSAM semantic information and enhancing feature descriptors, the model's matching accuracy and generalization capability are significantly improved.

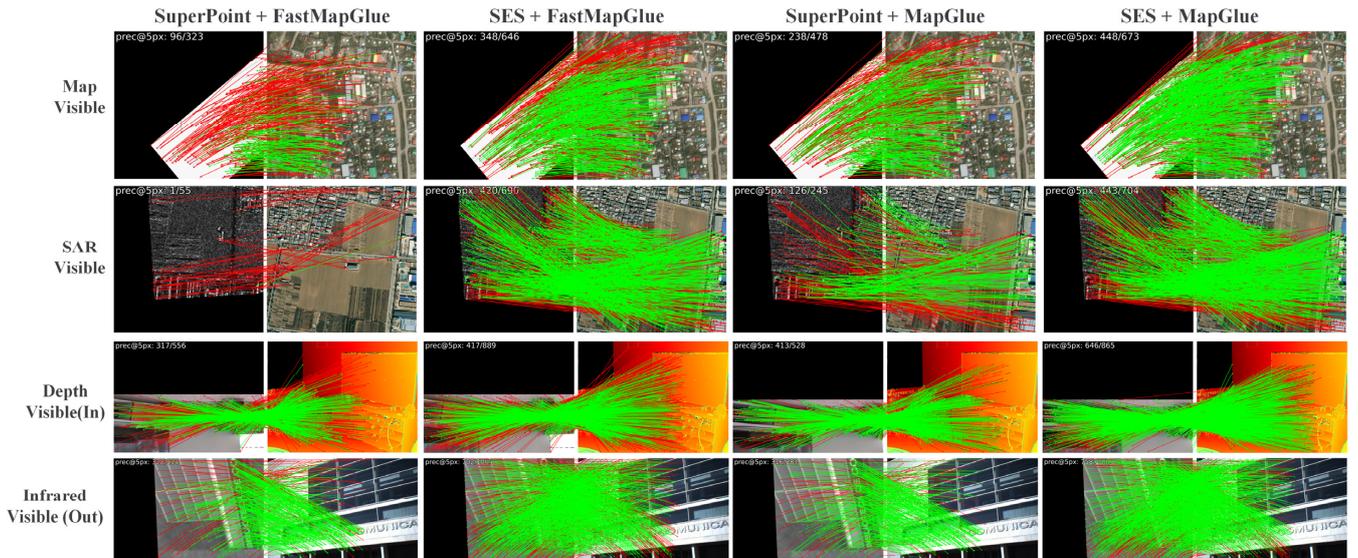

Fig 12. Qualitative Experimental Results Comparing SuperPoint and SES. (Green lines indicate projection errors below 5 pixels, while red lines indicate projection errors exceeding 5 pixels.)

To further validate the impact of the SES module, we evaluated the number of correct matching points within different error thresholds (3px, 5px, and 10px) on both the MapData-test-Hard and MultiResSAR-Low-Easy datasets, with the initial number of keypoints uniformly set to 2048. The effects of the SES module, compared to the original SuperPoint module, on the final number of correct matches are shown in Table VI. Under the @5px criterion, the FastMapGlue model achieves increases of 63.5% and 133.2% in the number of correct matches on the MapData-test-Hard and MultiResSAR-Low-Easy datasets, respectively, while the MapGlue model shows improvements of 62.3% and 139.4% under the same conditions. Qualitative comparison results are shown in Fig. 12. These results further indicate that the improved SES module can extract more robust keypoints and effectively reduce redundant and irrelevant features, thereby significantly enhancing the overall stability of the model.

## VI. CONCLUSION

In this work, we introduced MapData, a large-scale, globally covering, multi-scene, and multimodal dataset designed to address the challenging registration problem between electronic navigation maps and visible light images. MapData comprises 121,781 pairs of 512×512 pixels map–visible image pairs, each accompanied by ground-truth homography matrices generated through a combination of manual and automated annotation. The dataset's rich diversity and substantial sample size make it a valuable resource that can drive further research in MRSI matching. In addition, we proposed MapGlue, a cross-modal matching method that integrates semantic information using a dual graph neural network framework. By combining salient keypoint extraction, semantic enhancement, and dual graph structure guided matching enhancement, MapGlue achieves exceptional performance and generalization across multiple datasets, as demonstrated by our extensive experimental results.

While MapGlue exhibits excellent performance and stability in MRSI matching, its effectiveness is somewhat affected when handling natural images with large viewpoint differences. This limitation primarily arises because our method was trained exclusively on datasets containing map-visible image pairs, which offer very few cross-view samples. Consequently, the model has insufficient exposure to learning features that span significant viewpoint variations. To address this issue, future work can focus on two key directions: (1) incorporating training data with large viewpoint differences to enrich the range of view variations, and (2) designing specialized network architectures or loss functions that promote feature alignment across different viewpoints to further enhance generalization.

ACKNOWLEDGMENT

We gratefully acknowledge the substantial support of Yihang Cao, Zening Wang, Haojie Wang, Ziyi Hou, and others during the creation of this dataset. We also thank Google Maps for providing the rich data sources that made this work possible.

REFERENCES

[1] Z. Yongjun, Z. Zuxun, and G. Jianya, "Generalized photogrammetry of spaceborne, airborne and terrestrial multi-source remote sensing

datasets," Acta Geodaetica et Cartographica Sinica, vol. 50, no. 1, p. 1, 2021.
[2] N. G. Aditya, P. B. Dhruval, J. Shalabi, S. Jape, X. Wang, and Z. Jacob, "Thermal voyager: A comparative study of rgb and thermal cameras for night-time autonomous navigation," in 2024 IEEE International Conference on Robotics and Automation (ICRA), IEEE, 2024, pp. 14116–14122.
[3] Y. Ye, C. Yang, G. Gong, P. Yang, D. Quan, and J. Li, "Robust optical and SAR image matching using attention-enhanced structural features," IEEE Transactions on Geoscience and Remote Sensing, vol. 62, pp. 1–12, 2024.
[4] W. Yang, Y. Yao, Y. Zhang, and Y. Wan, "Weak texture remote sensing image matching based on hybrid domain features and adaptive description method," The Photogrammetric Record, vol. 38, no. 184, pp. 537–562, 2023.
[5] Y. Zhang et al., "Multi-modal remote sensing image robust matching based on Second-order tensor orientation feature transformation," IEEE Transactions on Geoscience and Remote Sensing, 2025.
[6] X. He et al., "MatchAnything: Universal Cross-Modality Image Matching with Large-Scale Pre-Training," arXiv preprint arXiv:2501.07556, 2025.
[7] Zhu Bai, Ye Yuanxin. 2024. Multimodal remote sensing image registration: a survey. Journal of Image and Graphics, 29(08):2137-2161 DOI: 10.11834/jig.230737
[8] Z. Li and N. Snavely, "Megadepth: Learning single-view depth prediction from internet photos," in Proceedings of the IEEE conference on computer vision and pattern recognition, 2018, pp. 2041–2050.
[9] J. L. Schonberger and J.-M. Frahm, "Structure-from-motion revisited," in Proceedings of the IEEE conference on computer vision and pattern recognition, 2016, pp. 4104–4113.
[10] A. Dai, A. X. Chang, M. Savva, M. Halber, T. Funkhouser, and M. Nießner, "Scannet: Richly-annotated 3d reconstructions of indoor scenes," in Proceedings of the IEEE conference on computer vision and pattern recognition, 2017, pp. 5828–5839.
[11] V. Balntas, K. Lenc, A. Vedaldi, and K. Mikolajczyk, "HPatches: A benchmark and evaluation of handcrafted and learned local descriptors," in Proceedings of the IEEE conference on computer vision and pattern recognition, 2017, pp. 5173–5182.
[12] J. L. Schonberger and J.-M. Frahm, "Structure-from-motion revisited," in Proceedings of the IEEE conference on computer vision and pattern recognition, 2016, pp. 4104–4113.
[13] C. A. Aguilera, A. D. Sappa, and R. Toledo, "LGHD: A feature descriptor for matching across non-linear intensity variations," in 2015 IEEE International Conference on Image Processing (ICIP), IEEE, 2015, pp. 178–181.
[14] X. Jia, C. Zhu, M. Li, W. Tang, and W. Zhou, "LLVIP: A visible-infrared paired dataset for low-light vision," in Proceedings of the IEEE/CVF international conference on computer vision, 2021, pp. 3496–3504.
[15] S. Hwang, J. Park, N. Kim, Y. Choi, and I. So Kweon, "Multispectral pedestrian detection: Benchmark dataset and baseline," in Proceedings of the IEEE conference on computer vision and pattern recognition, 2015, pp. 1037–1045.
[16] M. Brown and S. Süsstrunk, "Multi-spectral SIFT for scene category recognition," in CVPR 2011, IEEE, 2011, pp. 177–184.
[17] I. Vasiljevic et al., "Diode: A dense indoor and outdoor depth dataset," arXiv preprint arXiv:1908.00463, 2019.
[18] Ö. Tuzcuoğlu, A. Köksal, B. Sofu, S. Kalkan, and A. A. Alatan, "Xoftr: Cross-modal feature matching transformer," in Proceedings of the IEEE/CVF Conference on Computer Vision and Pattern Recognition, 2024, pp. 4275–4286.
[19] J. Li, Q. Hu, and Y. Zhang, "Multimodal image matching: A scale-invariant algorithm and an open dataset," ISPRS Journal of Photogrammetry and Remote Sensing, vol. 204, pp. 77–88, 2023.
[20] Y. Yao, Y. Zhang, Y. Wan, X. Liu, X. Yan, and J. Li, "Multi-modal remote sensing image matching considering co-occurrence filter," IEEE Transactions on Image Processing, vol. 31, pp. 2584–2597, 2022.
[21] Y. Zhao, X. Huang, and Z. Zhang, "Deep lucas-kanade homography for multimodal image alignment," in Proceedings of the IEEE/CVF conference on computer vision and pattern recognition, 2021, pp. 15950–15959.
[22] W. Zhang et al., "Multi-Resolution SAR and Optical Remote Sensing Image Registration Methods: A Review, Datasets, and Future Perspectives," arXiv preprint arXiv:2502.01002, 2025.
[23] D. G. Lowe, "Distinctive image features from scale-invariant keypoints," International journal of computer vision, vol. 60, pp. 91–110, 2004.
[24] H. Bay, T. Tuytelaars, and L. Van Gool, "Surf: Speeded up robust features," in Computer Vision–ECCV 2006: 9th European Conference on Computer Vision, Graz, Austria, May 7-13, 2006. Proceedings, Part I 9, Springer, 2006, pp. 404–417.
[25] W. Ma et al., "Remote sensing image registration with modified SIFT and enhanced feature matching," IEEE Geoscience and Remote Sensing Letters, vol. 14, no. 1, pp. 3–7, 2016.
[26] X. Xiong, G. Jin, Q. Xu, and H. Zhang, "Self-similarity features for multimodal remote sensing image matching," IEEE Journal of selected topics in applied earth Observations and remote sensing, vol. 14, pp. 12440–12454, 2021.
[27] X. Xiong, G. Jin, Q. Xu, H. Zhang, L. Wang, and K. Wu, "Robust registration algorithm for optical and SAR images based on adjacent self-similarity feature," IEEE Transactions on Geoscience and Remote Sensing, vol. 60, pp. 1–17, 2022.
[28] J. Li, Q. Hu, and M. Ai, "RIFT: Multi-modal image matching based on radiation-variation insensitive feature transform," IEEE Transactions on Image Processing, vol. 29, pp. 3296–3310, 2019.
[29] J. Li, P. Shi, Q. Hu, and Y. Zhang, "RIFT2: Speeding-up RIFT with a new rotation-invariance technique," arXiv preprint arXiv:2303.00319, 2023.
[30] Y. Yao, Y. Zhang, Y. Wan, X. Liu, and H. Guo, "Heterologous images matching considering anisotropic weighted moment and absolute phase orientation," Geomatics and Information Science of Wuhan University, vol. 46, no. 11, pp. 1727–1736, 2021.
[31] Y. Yao, B. Zhang, Y. Wan, and Y. Zhang, "MOTIF: Multi-orientation tensor index feature descriptor for SAR-optical image registration," The International Archives of the Photogrammetry, Remote Sensing and Spatial Information Sciences, vol. 43, pp. 99–105, 2022.
[32] Y. Zhang et al., "Histogram of the orientation of the weighted phase descriptor for multi-modal remote sensing image matching," ISPRS Journal of Photogrammetry and Remote Sensing, vol. 196, pp. 1–15, 2023.
[33] Y. Liao et al., "Refining multi-modal remote sensing image matching with repetitive feature optimization," International Journal of Applied Earth Observation and Geoinformation, vol. 134, p. 104186, 2024.
[34] Y. Liao, P. Tao, Q. Chen, L. Wang, and T. Ke, "Highly adaptive multi-modal image matching based on tuning-free filtering and enhanced sketch features," Information Fusion, vol. 112, p. 102599, 2024.
[35] Z. Hou, Y. Liu, and L. Zhang, "POS-GIFT: A geometric and intensity-invariant feature transformation for multimodal images," Information Fusion, vol. 102, p. 102027, 2024.
[36] D. DeTone, T. Malisiewicz, and A. Rabinovich, "Toward geometric deep slam," arXiv preprint arXiv:1707.07410, 2017.
[37] D. DeTone, T. Malisiewicz, and A. Rabinovich, "Superpoint: Self-supervised interest point detection and description," in Proceedings of the IEEE conference on computer vision and pattern recognition workshops, 2018, pp. 224–236.
[38] J. Revaud, C. De Souza, M. Humenberger, and P. Weinzaepfel, "R2d2: Reliable and repeatable detector and descriptor," Advances in neural information processing systems, vol. 32, 2019.
[39] M. Dusmanu et al., "D2-net: A trainable cnn for joint description and detection of local features," in Proceedings of the ieee/cvf conference on computer vision and pattern recognition, 2019, pp. 8092–8101.
[40] P.-E. Sarlin, D. DeTone, T. Malisiewicz, and A. Rabinovich, "Superglue: Learning feature matching with graph neural networks," in Proceedings of the IEEE/CVF conference on computer vision and pattern recognition, 2020, pp. 4938–4947.
[41] P. Lindenberger, P.-E. Sarlin, and M. Pollefeys, "Lightglue: Local feature matching at light speed," in Proceedings of the IEEE/CVF International Conference on Computer Vision, 2023, pp. 17627–17638.
[42] H. Jiang, A. Karpur, B. Cao, Q. Huang, and A. Araujo, "Omniglue: Generalizable feature matching with foundation model guidance," in Proceedings of the IEEE/CVF Conference on Computer Vision and Pattern Recognition, 2024, pp. 19865–19875.

[43] M. Oquab et al., "Dinov2: Learning robust visual features without supervision," arXiv preprint arXiv:2304.07193, 2023.
[44] G. Potje, F. Cadar, A. Araujo, R. Martins, and E. R. Nascimento, "Xfeat: Accelerated features for lightweight image matching," in Proceedings of the IEEE/CVF Conference on Computer Vision and Pattern Recognition, 2024, pp. 2682–2691.
[45] Y. Gao, J. He, T. Zhang, Z. Zhang, and Y. Zhang, "Dynamic keypoint detection network for image matching," IEEE Transactions on Pattern Analysis and Machine Intelligence, vol. 45, no. 12, pp. 14404–14419, 2023.
[46] J. Sun, Z. Shen, Y. Wang, H. Bao, and X. Zhou, "LoFTR: Detector-free local feature matching with transformers," in Proceedings of the IEEE/CVF conference on computer vision and pattern recognition, 2021, pp. 8922–8931.
[47] Y. Wang, X. He, S. Peng, D. Tan, and X. Zhou, "Efficient LoFTR: Semi-dense local feature matching with sparse-like speed," in Proceedings of the IEEE/CVF conference on computer vision and pattern recognition, 2024, pp. 21666–21675.
[48] K. He, X. Chen, S. Xie, Y. Li, P. Dollár, and R. Girshick, "Masked autoencoders are scalable vision learners," in Proceedings of the IEEE/CVF conference on computer vision and pattern recognition, 2022, pp. 16000–16009.
[49] J. Edstedt, I. Athanasiadis, M. Wadenbäck, and M. Felsberg, "DKM: Dense kernelized feature matching for geometry estimation," in Proceedings of the IEEE/CVF Conference on Computer Vision and Pattern Recognition, 2023, pp. 17765–17775.
[50] J. Edstedt, Q. Sun, G. Bökman, M. Wadenbäck, and M. Felsberg, "RoMa: Robust dense feature matching," in Proceedings of the IEEE/CVF Conference on Computer Vision and Pattern Recognition, 2024, pp. 19790–19800.
[51] H. Zhu et al., "Mcnet: Rethinking the core ingredients for accurate and efficient homography estimation," in Proceedings of the IEEE/CVF Conference on Computer Vision and Pattern Recognition, 2024, pp. 25932–25941.
[52] J. Liu, X. Li, Q. Wei, J. Xu, and D. Ding, "Semi-supervised keypoint detector and descriptor for retinal image matching," in European Conference on Computer Vision, Springer, 2022, pp. 593–609.
[53] X. Shen et al., "GIM: Learning Generalizable Image Matcher From Internet Videos," presented at the ICLR, Jan. 2024. Accessed: Mar. 18, 2025.
[54] X. Jiang, J. Ren, Z. Li, X. Zhou, D. Liang, and X. Bai, "MINIMA: Modality Invariant Image Matching," arXiv preprint arXiv:2412.19412, 2024.
[55] L. Moisan and B. Stival, "A probabilistic criterion to detect rigid point matches between two images and estimate the fundamental matrix," International Journal of Computer Vision, vol. 57, pp. 201–218, 2004.
[56] C. Zhang et al., "Faster segment anything: Towards lightweight sam for mobile applications," arXiv preprint arXiv:2306.14289, 2023.
[57] A. Vaswani et al., "Attention is all you need," Advances in neural information processing systems, vol. 30, 2017.
[58] A. Kirillov et al., "Segment anything," in Proceedings of the IEEE/CVF international conference on computer vision, 2023, pp. 4015–4026.
[59] Z. Ye et al., "Comparison and evaluation of feature matching methods for multisource planetary remote sensing imagery", Photogramm. Rec., vol. 39, no. 188, pp. 845-875, Dec. 2024.
[60] S. Ji, C. Zeng, Y. Zhang, and Y. Duan, "An evaluation of conventional and deep learning‐based image‐matching methods on diverse datasets," The Photogrammetric Record, vol. 38, no. 182, pp. 137–159, 2023.